\documentclass[10pt]{article} 
\usepackage[accepted]{tmlr}

\usepackage{amsmath,amsfonts,bm}

\def\eqref#1{equation~\ref{#1}}

\def\1{\bm{1}}

\DeclareMathAlphabet{\mathsfit}{\encodingdefault}{\sfdefault}{m}{sl}
\SetMathAlphabet{\mathsfit}{bold}{\encodingdefault}{\sfdefault}{bx}{n}

\usepackage{hyperref}
\usepackage{url}

\usepackage{booktabs}
\usepackage{enumitem}
\setlist[itemize]{leftmargin=*}

\usepackage{graphicx}
\usepackage{amsmath}
\usepackage{wrapfig}
\usepackage{array}
\usepackage{multirow}
\usepackage{makecell}
\usepackage{caption}
\usepackage{subcaption}
\usepackage{bbm}

\def\model{G4SATBench}

\title{\model{}: Benchmarking and Advancing SAT Solving with Graph Neural Networks}

\author{\name Zhaoyu Li \email zhaoyu@cs.toronto.edu \\
        \addr University of Toronto
        \AND
        \name Jinpei Guo \email mike0728@sjtu.edu.cn \\
        \addr Shanghai Jiao Tong University
        \AND
        \name Xujie Si \email six@cs.toronto.edu \\
        \addr University of Toronto \\
}

\def\month{05}  
\def\year{2024} 
\def\openreview{\url{https://openreview.net/forum?id=7VB5db72lr}} 

\begin{document}

\maketitle

\begin{abstract}
Graph neural networks (GNNs) have recently emerged as a promising approach for solving the Boolean Satisfiability Problem (SAT), offering potential alternatives to traditional backtracking or local search SAT solvers. However, despite the growing volume of literature in this field, there remains a notable absence of a unified dataset and a fair benchmark to evaluate and compare existing approaches. To address this crucial gap, we present \model{}, the first benchmark study that establishes a comprehensive evaluation framework for GNN-based SAT solvers. In \model{}, we meticulously curate a large and diverse set of SAT datasets comprising 7 problems with 3 difficulty levels and benchmark a broad range of GNN models across various prediction tasks, training objectives, and inference algorithms. To explore the learning abilities and comprehend the strengths and limitations of GNN-based SAT solvers, we also compare their solving processes with the heuristics in search-based SAT solvers. Our empirical results provide valuable insights into the performance of GNN-based SAT solvers and further suggest that existing GNN models can effectively learn a solving strategy akin to greedy local search but struggle to learn backtracking search in the latent space. Our codebase is available at \url{https://github.com/zhaoyu-li/G4SATBench}.
\end{abstract}

\section{Introduction}
\label{sec:introduction}
The Boolean Satisfiability Problem (SAT) is a crucial problem at the nexus of computer science, logic, and operations research, which has garnered significant attention over the past five decades. To solve SAT instances efficiently, modern SAT solvers have been developed with backtracking (especially with conflict-driven clause learning, a.k.a. CDCL) or local search (LS) heuristics that effectively exploit the instance's structure and traverse its vast search space~\citep{biere2009handbook}. However, designing such heuristics remains a highly non-trivial and time-consuming task, with a lack of significant improvement in recent years. Conversely, the recent rapid advances in graph neural networks (GNNs)~\citep{ggnn,gcn,gin} have shown impressive performances in analyzing structured data, offering a promising opportunity to enhance or even replace modern SAT solvers. As such, there have been massive efforts to leverage GNNs to solve SAT over the last few years~\citep{holden2021machine,guo2022machine}.

Despite the recent progress, the question of \textit{how (well) GNNs can solve SAT} remains unanswered. One of the main reasons for this is the variety of learning objectives and usage scenarios employed in existing work, making it difficult to evaluate different methods in a fair and comprehensive manner. For example, NeuroSAT~\citep{neurosat} predicts satisfiability, NeuroCore~\citep{neurocore} classifies unsat-core variables, QuerySAT~\citep{querysat} constructs a satisfying assignment, and NSNet~\citep{nsnet} predicts marginal distributions of all satisfying solutions to solve the SAT problem. Moreover, most previous research has experimented on different datasets that vary in a range of settings (e.g., data distribution, instance size, and dataset size), which leads to a lack of unified and standardized datasets for training and evaluation. Additionally, some work~\citep{pdp, satformer, addressing} has noted the difficulty of re-implementing prior approaches as baselines, rendering it arduous to draw consistent conclusions about the performance of peer methods. All of these issues impede the development of GNN-based solvers for SAT solving. 

To systematically quantify the progress in this field and facilitate rapid, reproducible, and generalizable research, we propose \textbf{\model{}}, the first comprehensive benchmark study for SAT solving with GNNs. \model{} is characterized as follows:

\begin{itemize}
\item First, we construct a large and diverse collection of SAT datasets that includes instances from distinct sources and difficulty levels. Specifically, our benchmark consists of 7 different datasets from 3 benchmark families, including random instances, pseudo-industrial instances, and combinatorial problems. It not only covers a wide range of prior datasets but also introduces 3 levels of difficulty for each dataset to enable fine-grained analyses.

\item Second, we re-implement various GNN-based SAT solvers with unified interfaces and configuration settings, establishing a general evaluation protocol for fair and comprehensive comparisons. Our framework allows for evaluating different GNN models in SAT solving with various prediction tasks, training objectives, and inference algorithms, encompassing the diverse learning frameworks employed in the existing literature.

\item Third, we present baseline results and conduct thorough analyses of GNN-based SAT solvers, providing a detailed reference of prior work and laying a solid foundation for future research. Our evaluations assess the performances of different choices of GNN models (e.g., graph constructions, message-passing schemes) with particular attention to some critical parameters (e.g., message-passing iterations), as well as their generalization ability across different distributions.

\item Lastly, we conduct a series of in-depth experiments to explore the learning abilities of GNN-based SAT solvers. Specifically, we compare the training and solving processes of GNNs with the heuristics employed in both CDCL and LS-based SAT solvers. Our experimental results reveal that \textit{GNNs tend to develop a solving heuristic similar to greedy local search to find a satisfying assignment but fail to effectively learn the CDCL heuristic in the latent space.}
\end{itemize}

We believe that \model{} will help the research community to make significant strides in understanding the capabilities and limitations of GNNs for solving SAT and facilitate further endeavors in this domain.

\section{Related Work}
\paragraph{SAT solving with GNNs.}
Existing GNN-based SAT solvers can be broadly categorized into two branches~\citep{holden2021machine,guo2022machine}: \textit{standalone neural solvers} and \textit{neural-guided solvers}. Standalone neural solvers utilize GNNs to solve SAT instances directly. For example, a stream of research~\citep{bunz2017graph, neurosat, jaszczur2020neural, cameron2020predicting, satformer} focuses on predicting the satisfiability of a given formula, while several alternative approaches~\citep{circuitsat, pdp, querysat, deepsat, addressing} aim to construct a satisfying assignment. Neural-guided solvers, on the other hand, integrate GNNs with modern SAT solvers, trying to improve their search heuristics with the prediction of GNNs. These methods typically train GNN models using supervised learning on some tasks such as unsat-core variable prediction~\citep{neurocore, neurocomb}, satisfying assignment prediction~\citep{nlocalsat}, glue variable prediction~\citep{neuroglue}, and assignment marginal prediction~\citep{nsnet}, or through reinforcement learning~\citep{rllocalsearch, rlcdcl} by modeling the entire search procedure as a Markov decision process. Despite the rich literature on SAT solving with GNNs, there is no benchmark study to evaluate and compare the performance of these GNN models. We hope the proposed \model{} would address this gap.

\paragraph{SAT datasets.}
Several established SAT benchmarks, including the prestigious SATLIB~\citep{satlib} and the SAT Competitions over the years, have provided a variety of practical instances to assess the performance of modern SAT solvers. Regrettably, these datasets are not particularly amenable for GNNs to learn from, given their relatively modest scale (less than 100 instances for a specific domain) or overly extensive instances (exceeding 10 million variables and clauses). To address this issue, researchers have turned to synthetic SAT instance generators~\citep{ca, ps, cnfgen, neurosat}, which allow for the creation of a flexible number of instances with customizable settings. However, most of the existing datasets generated from these sources are limited to a few domains (less than 3 generators), small in size (less than 10k instances), or easy in difficulty (less than 40 variables within an instance), and there is no standardized dataset for evaluation. In \model{}, we include a variety of synthetic generators with carefully selected configurations, aiming to construct a broad collection of SAT datasets that are highly conducive for training and evaluating GNNs.

\section{Preliminaries}
\paragraph{The SAT problem.}
In propositional logic, a Boolean formula is constructed from Boolean variables and logical operators such as conjunctions ($\land$), disjunctions ($\lor$), and negations ($\neg$). It is typical to represent Boolean formulas in conjunctive normal form (CNF), expressed as a conjunction of clauses, where each clause is a disjunction of literals, which can be either a variable or its negation. Given a CNF formula, the SAT problem is to determine if there exists an assignment of boolean values to its variables such that the formula evaluates to true. If this is the case, the formula is called satisfiable; otherwise, it is unsatisfiable. For a satisfiable instance, one is expected to construct a satisfying assignment to prove its satisfiability. On the other hand, for an unsatisfiable formula, one can find a minimal subset of clauses whose conjunction is still unsatisfiable. Such a set of clauses is termed the unsat core, and variables in the unsat core are referred to as unsat-core variables.

\paragraph{Graph representations of CNF formulas.}
Traditionally, a CNF formula can be represented using 4 types of graphs~\citep{biere2009handbook}: Literal-Clause Graph (LCG), Variable-Clause Graph (VCG), Literal-Incidence Graph (LIG), and Variable-Incidence Graph (VIG). The LCG is a bipartite graph with literal and clause nodes connected by edges indicating the presence of a literal in a clause. The VCG is formed by merging the positive and negative literals of the same variables in LCG. The LIG, on the other hand, only consists of literal nodes, with edges indicating co-occurrence in a clause. Lastly, the VIG is derived from LIG using the same merging operation as VCG.
\section{\model{}: A Benchmark Study on GNNs for SAT Solving}
\begin{figure}[t]
\begin{center}
\includegraphics[width=\columnwidth]{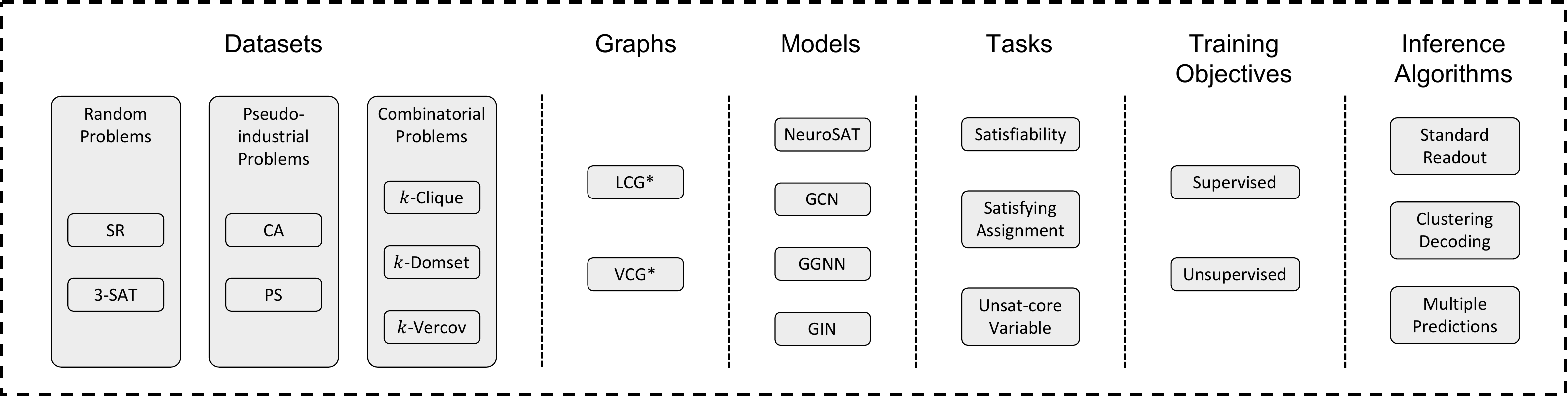}
\caption{Framework overview of \model{}.}
\label{fig:overview}
\end{center}
\end{figure}

The goal of \model{} is to establish a general framework that enables comprehensive comparisons and evaluations of various GNN-based SAT solvers. In this section, we will delve into the details of \model{}, including its datasets, GNN models, prediction tasks, as well as training and testing methodologies. The overview of the \model{} framework is shown in Figure~\ref{fig:overview}.

\subsection{Datasets}
\model{} is built on a diverse set of synthetic CNF generators. It currently consists of 7 datasets sourced from 3 distinct domain areas: random problems, pseudo-industrial problems, and combinatorial problems. Specifically, we utilize the SR generator in NeuroSAT~\citep{neurosat} and the 3-SAT generator in CNFGen~\citep{cnfgen} to produce random CNF formulas. For pseudo-industrial problems, we employ the Community Attachment (CA) model~\citep{ca} and the Popularity-Similarity (PS) model~\citep{ps}, which generate synthetic instances that exhibit similar statistical features, such as the community and the locality, to those observed in real-world industrial SAT instances. For combinatorics, we resort to 3 synthetic generators in CNFGen~\citep{cnfgen} to create SAT instances derived from the translation of $k$-Clique, $k$-Dominating Set, and $k$-Vertex Cover problems.

In addition to the diversity of datasets, \model{} offers distinct difficulty levels for all datasets to enable fine-grained analyses. These levels include easy, medium, and hard, with the latter representing more complex problems with increased instance sizes. For example, the easy SR dataset contains instances with 10 to 40 variables, the medium SR dataset contains formulas with 40 to 200 variables, and the hard SR dataset consists of formulas with variables ranging from 200 to 400. For each easy and medium dataset, we generate 80k pairs of satisfiable and unsatisfiable instances for training, 10k pairs for validation, and 10k pairs for testing. For each hard dataset, we produce 10k testing pairs. It is also worth noting that the parameters for our synthetic generators are meticulously selected to avoid generating trivial cases. For instance, we produce random 3-SAT formulas at the phase-transition region where the relationship between the number of clauses $(m)$ and variables $(n)$ is $m = 4.258n +58.26n^{-2/3}$~\citep{3-sat}, and utilize the $v$ vertex Erd\H{o}s-R\'{e}nyi graph with an edge probability of $p = \binom{v}{k}^{-1/\binom{v}{2}}$ to generate $k$-Clique problems, making the expected number of $k$-Cliques in a graph equals 1~\citep{k-cliques}. To provide a detailed characterization of our generated datasets, we compute several statistics of the SAT instances across difficulty levels in \model{}. Please refer to Appendix~\ref{app:dataset} for more information about the datasets.

\subsection{GNN Baselines}
\paragraph{Graph constructions.}
\begin{wrapfigure}{r}{0.45\textwidth}
\begin{center}
\vspace{-15pt}
\includegraphics[width=0.45\columnwidth]{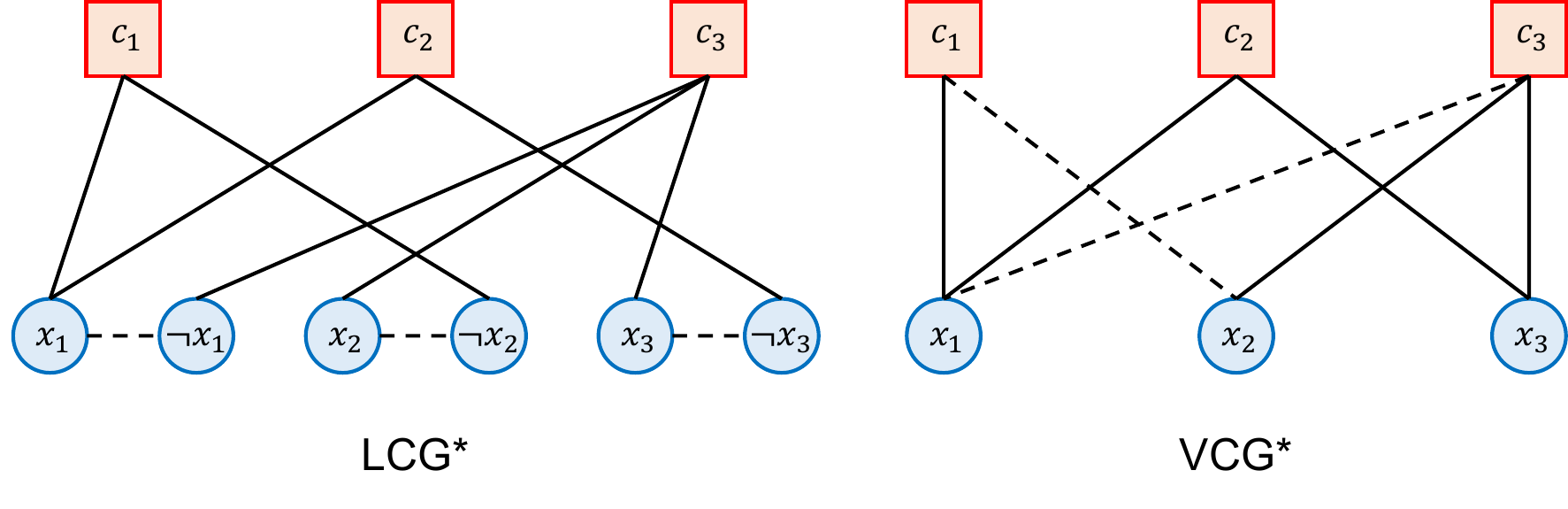}
\caption{LCG* and VCG* of the CNF formula $(x_1 \lor \neg x_2) \land (x_1 \lor x_3) \land(\neg x_1 \lor x_2 \lor x_3)$.}
\label{fig:lcg_vcg}
\vspace{-5pt}
\end{center}
\end{wrapfigure}
It is important to note that traditional graph representations of a CNF formula often lack the requisite details for optimally constructing GNNs. Specifically, the LIG and VIG exclude clause-specific information, while the LCG and VCG fail to differentiate between positive and negative literals of the same variable. To address these limitations, existing approaches typically build GNN models on the refined versions of the LCG and VCG encodings. In the LCG, a new type of edge is added between each literal and its negation, while the VCG is modified by using two types of edges to indicate the polarities of variables within a clause. These modified encodings are termed the LCG* and VCG* respectively, and an example of them is shown in Figure~\ref{fig:lcg_vcg}. It is also worth noting alternative graph encodings like the And-Inverter-Graph (AIG), can be applied for SAT instances that are not in CNF. However, such representations are specialized to specific applications (like CircuitSAT) and are not designed for general purposes. Given this specialization, we choose to keep them outside the scope of the current G4SATBench.

\paragraph{Message-passing schemes.}
\model{} enables performing various \textit{hetergeneous} message-passage algorithms between neighboring nodes on the LCG* or VCG* encodings of a CNF formula. For the sake of illustration, we will take GNN models on the LCG* as an example. We first define a $d$-dimensional embedding for every literal node and clause node, denoted by $h_l$ and $h_c$ respectively. Initially, all these embeddings are assigned to two learnable vectors $h_l^0$ and $h_c^0$, depending on their node types. At the $k$-th iteration of message passing, these hidden representations are updated as:
\begin{equation}
\label{eqn:gnn_lcg}
\begin{aligned}
h_c^{(k)} &= \text{UPD}\left(\underset{l \in \mathcal{N}(c)}{\text{AGG}}\left(\left\{\text{MLP}\left(h_l^{(k-1)}\right)\right\}\right), h_c^{(k-1)}\right), \\
h_l^{(k)} &= \text{UPD}\left(\underset{c \in \mathcal{N}(l)}{\text{AGG}}\left(\left\{\text{MLP}\left(h_c^{(k-1)}\right)\right\}\right), h_{\neg l}^{(k-1)}, h_l^{(k-1)}\right),
\end{aligned}
\end{equation}
where $\mathcal{N}(\cdot)$ denotes the set of neighbor nodes, $\text{MLP}$ is the multi-layer perception, $\text{UPD}(\cdot)$ is the update function, and $\text{AGG}(\cdot)$ is the aggregation function. Most GNN models on LCG* use Equation~\ref{eqn:gnn_lcg} with different choices of the update function and aggregation function. For instance, NeuroSAT employs LayerNormLSTM~\citep{ba2016layer} as the update function and summation as the aggregation function. In \model{}, we provide a diverse range of GNN models, including NeuroSAT~\citep{neurosat}, Graph Convolutional Network~(GCN)~\citep{gcn}, Gated Graph Neural Network~(GGNN)~\citep{ggnn}, and Graph Isomorphism Network~(GIN)~\citep{gin}, on the both LCG* and VCG*. More details of these GNN models are included in Appendix~\ref{app:gnn}.

\subsection{Supported Tasks, Training and Testing Settings}
\paragraph{Prediction tasks.}
In \model{}, we support three essential prediction tasks for SAT solving: satisfiability prediction, satisfying assignment prediction, and unsat-core variable prediction. These tasks are widely used in both standalone neural solvers and neural-guided solvers. Technically, we model satisfiability prediction as a binary graph classification task, where 1/0 denotes the given SAT instance $\phi$ is satisfiable/unsatisfiable. Here, we take GNN models on the LCG* as an example. After $T$ message passing iterations, we obtain the graph embedding by applying mean pooling on all literal embeddings, and then predict the satisfiability using an MLP followed by the sigmoid function $\sigma$:
\begin{equation}
y_{\phi} = \sigma\left(\text{MLP}\left(\text{MEAN}\left(\{h_l^{(T)}, l \in \phi\}\right)\right)\right).
\end{equation}

For satisfying assignment prediction and unsat-core variable prediction, we formulate them as binary node classification tasks, predicting the label for each variable in the given CNF formula $\phi$. In the case of GNNs on the LCG*, we concatenate the embeddings of each pair of literals $h_l$ and $h_{\neg l}$ to construct the variable embedding, and then readout using an MLP and the sigmoid function $\sigma$:
\begin{equation}
y_{v} = \sigma\left(\text{MLP}\left(\left[h_l^{(T)}, h_{\neg l}^{(T)}\right]\right)\right).
\end{equation}

\paragraph{Training objectives.}
To train GNN models on the aforementioned tasks, one common approach is to minimize the binary cross-entropy loss between the predictions and the ground truth labels. In addition to supervised learning, \model{} supports two unsupervised training paradigms for satisfying assignment prediction~\citep{circuitsat, querysat}. The first approach aims to differentiate and maximize the satisfiability value of a CNF formula~\citep{circuitsat}. It replaces the $\neg$ operator with the function $N(x_i) = 1 - x_i$ and uses smooth max and min functions to replace the $\lor$ and $\land$ operators. The smooth max and min functions are defined as follows:
\begin{equation} \label{eq:smooth_operators}
S_{max} (x_1, x_2, \dots, x_d) = \frac{\sum_{i=1}^{d}x_i\cdot e^{x_i/\tau}}{\sum_{i=1}^{d}e^{x_i/\tau}}, \quad
S_{min} (x_1, x_2, \dots, x_d) = \frac{\sum_{i=1}^{d}x_i\cdot e^{-x_i/\tau}}{\sum_{i=1}^{d}e^{-x_i/\tau}}, 
\end{equation}
where $\tau \geq 0$ is the temperature parameter. Given a predicted assignment $x$, we apply the smoothing logical operators and substitute variables in a formula $\phi$ with the corresponding values from $x$ to calculate its satisfiability value $S(x)$. Then we can minimize the following loss function:
\begin{equation} \label{eq:unsup}
    \mathcal{L}_{\phi}(x) = \frac{\left(1 - S(x)\right)^{\kappa}}{\left(1 - S(x)\right)^{\kappa} + S(x)^{\kappa}}. \quad (\kappa \geq 1\ \text{is a predefined constant})
\end{equation}

The second unsupervised loss is defined as follows~\citep{querysat}:
\begin{equation} \label{eq:unsupv2}
    V_c(x) = 1 - \prod_{i\in c^{+}}(1-x_i)\prod_{i\in c^{-}}x_i, \quad
    \mathcal{L}_{\phi}(x) = -\log\Bigl(\prod_{c\in\phi}V_c(x)\Bigr) = - \sum_{c \in \phi} \log \left(V_c(x)\right),
\end{equation}
where $c^{+}$ and $c^{-}$ are the sets of variables that occur in the clause $c$ in positive and negative form respectively. Note that these two losses reach the minimum only when the prediction $x$ is a satisfying assignment, thus minimizing such losses could help to construct a possible satisfying assignment.

\paragraph{Inference algorithms.}
Beyond the standard readout process like training, \model{} offers two alternative inference algorithms for satisfying assignment prediction~\citep{neurosat, pdp}. The first method performs 2-clustering on the literal embeddings to obtain two centers $\Delta_1$ and $\Delta_2$ and then partitions the positive and negative literals of each variable into distinct groups based on the predicate $||x_i - \Delta_1||^2 + ||\neg x_i - \Delta_2 ||^2 < ||x_i - \Delta_2||^2 + ||\neg x_i - \Delta_1||^2$~\citep{neurosat}. This allows the construction of two possible assignments by mapping one group of literals to true. The second approach is to employ the readout function at each iteration of message passing, resulting in multiple assignment predictions for a given instance~\citep{pdp}.

\paragraph{Evaluation metrics.}
For satisfiability prediction and unsat-core variable prediction, we report the classification accuracy of each GNN model in \model{}. For satisfying assignment prediction, we report the solving accuracy of the predicted assignments. If multiple assignments are predicted for a SAT instance, the instance is considered solved if any of the predictions satisfy the formula.

\section{Benchmarking Evaluation on \model{}}
\label{sec:benchmark}
In this section, we present the benchmarking results of \model{}. To ensure a fair comparison, we conduct a grid search to tune the hyperparameters of each GNN baseline. The best checkpoint for each GNN model is selected based on its performance on the validation set. To mitigate the impact of randomness, we use 3 different random seeds to repeat the experiment in each setting and report the average performance. Each experiment is performed on a single RTX8000 GPU and 16 AMD EPYC 7502 CPU cores, and the total time cost is approximately 8,000 GPU hours. For detailed experimental setup and hyperparameters, please refer to Appendix~\ref{app:ben_imp}.

\subsection{Satisfiability Prediction}
\paragraph{Evaluation on the same distribution.}
Table~\ref{tab:satisfiability_general} shows the benchmarking results of each GNN baseline when trained and evaluated on datasets possessing identical distributions. All GNN models exhibit strong performance across most easy and medium datasets, except for the medium SR dataset. This difficulty can be attributed to the inherent characteristic of this dataset, which includes satisfiable and unsatisfiable pairs of medium-sized instances distinguished by just a single differing literal. Such a subtle difference presents a substantial challenge for GNN models in satisfiability classification. Among all GNN models, the different graph constructions do not seem to have a significant impact on the results, and NeuroSAT (on LCG*) and GGNN (on VCG*) achieve the best overall performance.

\begin{table}[ht]
  \caption{Classification accuracy of satisfiability on identical distribution.}
  \label{tab:satisfiability_general}
  \centering
  \resizebox{\textwidth}{!}{
  \begin{tabular}{llcccccccccccccc}
    \toprule[1pt]
    \multirow{2}{*}{\bf Graph} & \multirow{2}{*}{\bf Method} & \multicolumn{7}{c}{\bf Easy Datasets} & \multicolumn{7}{c}{\bf Medium Datasets} \\
    \cmidrule(lr){3-9}
    \cmidrule(lr){10-16}
     & & \bf SR & \bf 3-SAT & \bf CA & \bf PS & \bf $k$-Clique & \bf $k$-Domset & \bf $k$-Vercov & \bf SR & \bf 3-SAT & \bf CA & \bf PS & \bf $k$-Clique & \bf $k$-Domset & \bf $k$-Vercov \\
    \midrule
    \multirow{4}{*}{LCG*} & NeuroSAT & 96.00 & \textbf{96.33} & \textbf{98.83} & 96.59 & 97.92 & \textbf{99.77} & \textbf{99.99} & \textbf{78.02} & \textbf{84.90} & \textbf{99.57} & \textbf{96.81} & \textbf{89.39} & 99.67 & 99.80 \\
    & GCN & 94.43 & 94.47 & 98.79 & \textbf{97.53} & 98.24 & 99.59 & 99.98 & 69.39 & 82.67 & 99.53 & 96.16 & 85.72 & 99.16 & 99.74 \\
    & GGNN & \textbf{96.36} & 95.70 & 98.81 & 97.47 & \textbf{98.80} & 99.77 & 99.97 & 71.44 & 83.45 & 99.50 & 96.21 & 81.20 & \textbf{99.69} & \textbf{99.83} \\
    & GIN & 95.78 & 95.37 & 98.14 & 96.98 & 97.60 & 99.71 & 99.97 & 70.54 & 82.80 & 99.49 & 95.80 & 83.87 & 99.61 & 99.62\\
    \midrule
    \multirow{3}{*}{VCG*} & GCN & 93.19 & 94.92 & 97.82 & 95.79 & 98.72 & 99.54 & \textbf{99.99} & 66.35 & 83.75 & 99.49 & 95.48 & 82.99 & 99.42 & \textbf{99.89} \\
    & GGNN & \textbf{96.75} & \textbf{96.25} & \textbf{98.7}7 & 96.44 & \textbf{98.88} & \textbf{99.68} & 99.98 & \textbf{77.12} & 85.11 & \textbf{99.57} & 96.48 & 83.63 & \textbf{99.62} & 98.92 \\
     & GIN & 96.04 & 95.71 & 98.47 & \textbf{96.95} & 97.33 & 99.59 & 99.98 & 73.56 & \textbf{85.26} & 99.49 & \textbf{96.55} & \textbf{89.41} & 99.38 & 99.80 \\
    \bottomrule[1pt]
  \end{tabular}
  }
\end{table}

\begin{figure*}[ht]
    \centering
    \resizebox{\textwidth}{!}{
        \includegraphics{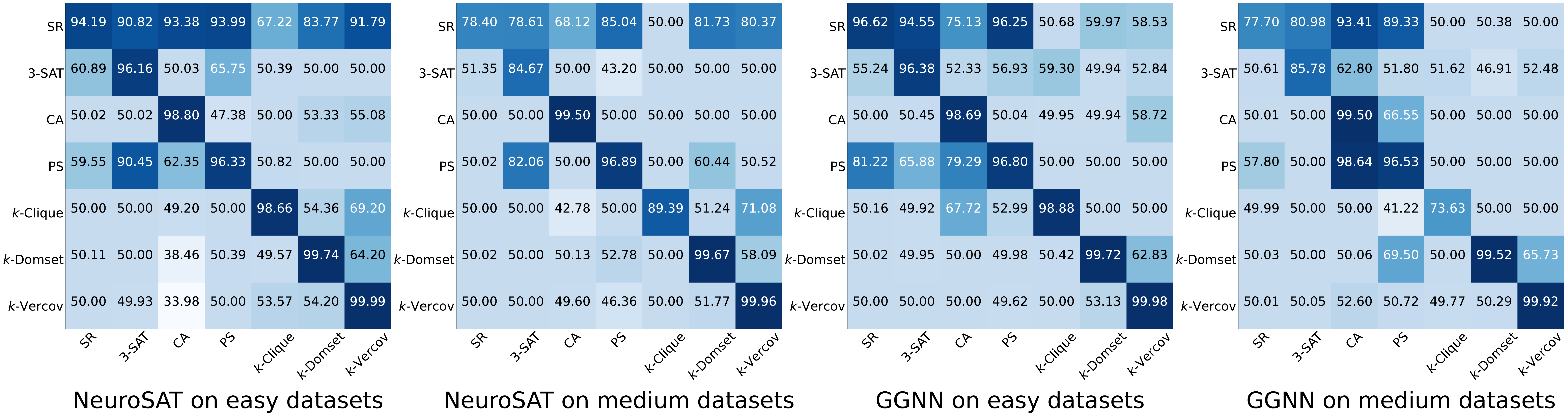}
    }
    \caption{Classification accuracy of satisfiability across different datasets. The x-axis denotes testing datasets and the y-axis denotes training datasets.}
    \label{fig:sat-dataset}
\end{figure*}

\begin{figure*}[ht]
    \begin{center}
    \resizebox{\textwidth}{!}{
        \begin{tabular}{cccc}
            \includegraphics[width=0.24\textwidth]{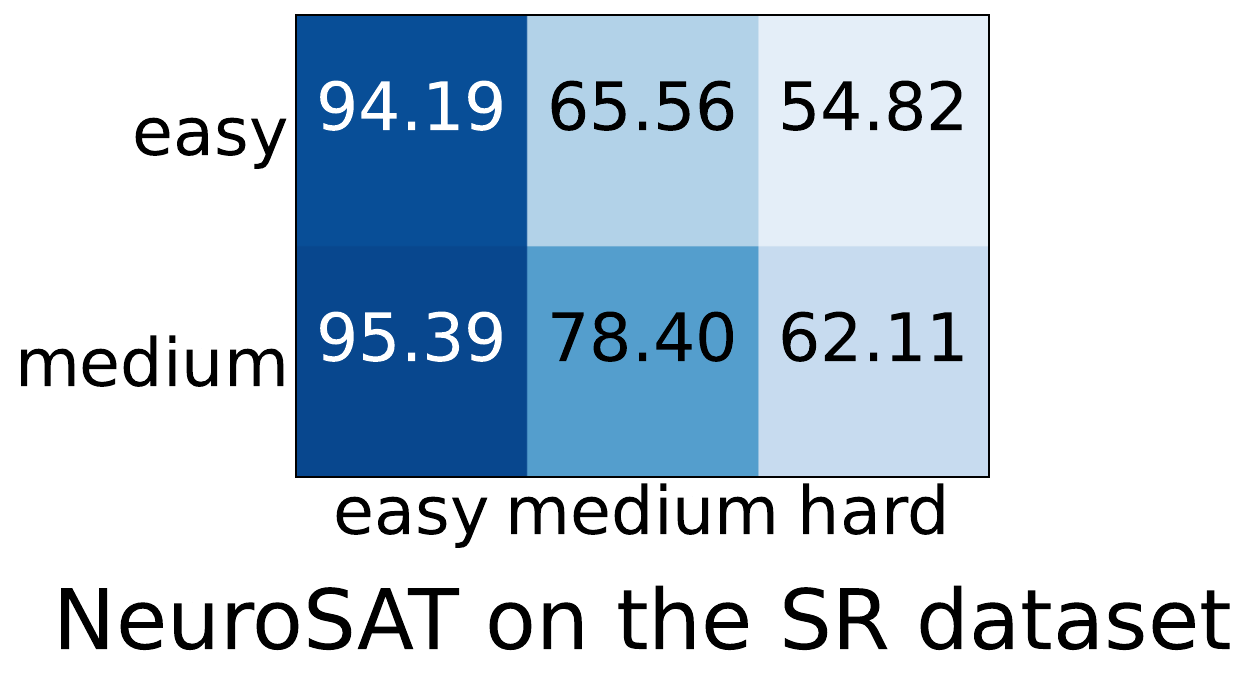}  &
            \includegraphics[width=0.255\textwidth]{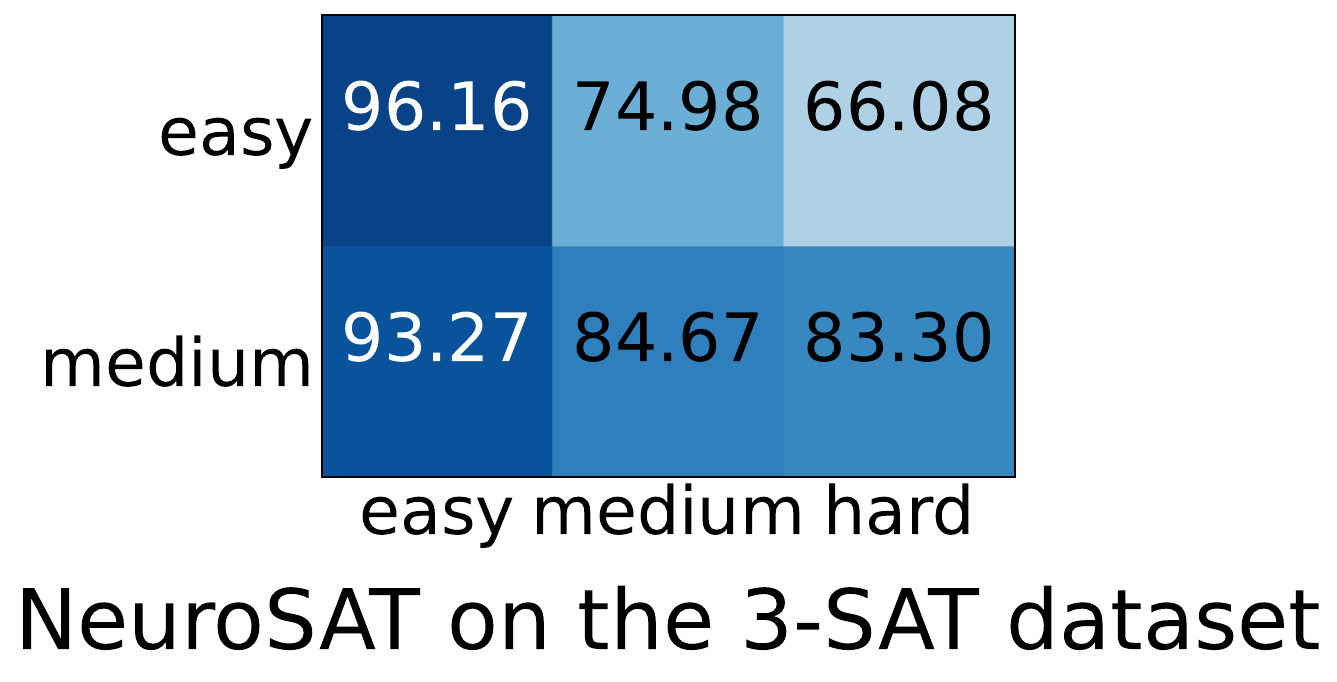} &
            \includegraphics[width=0.223\textwidth]{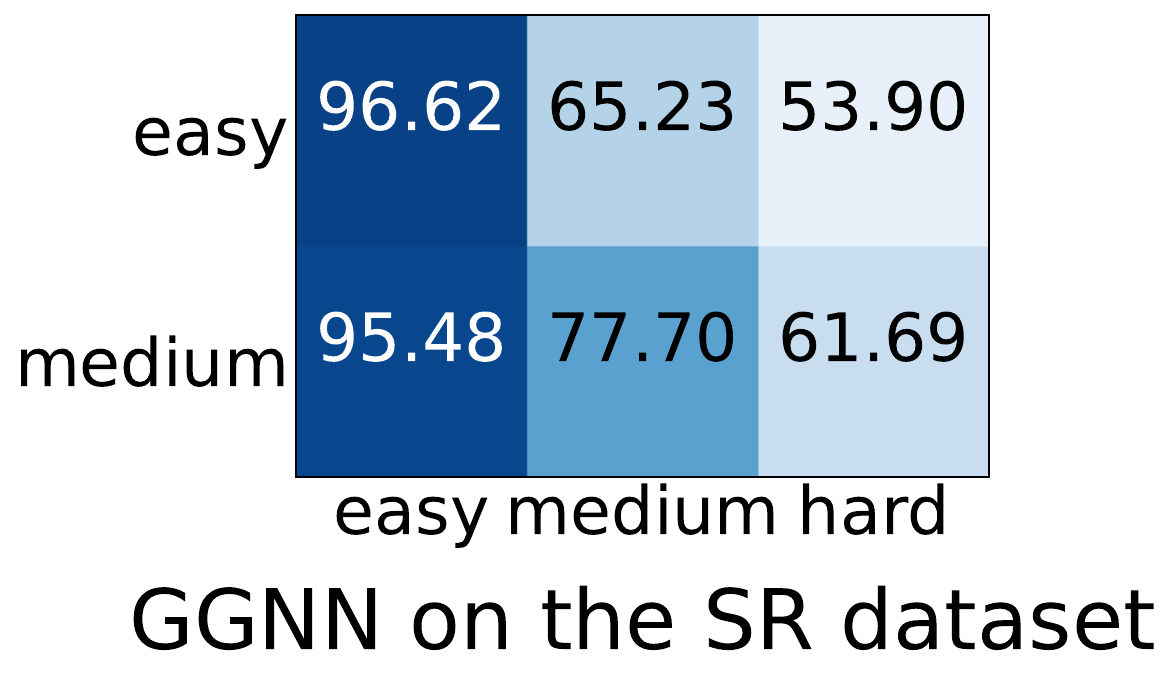} &
            \includegraphics[width=0.235\textwidth]{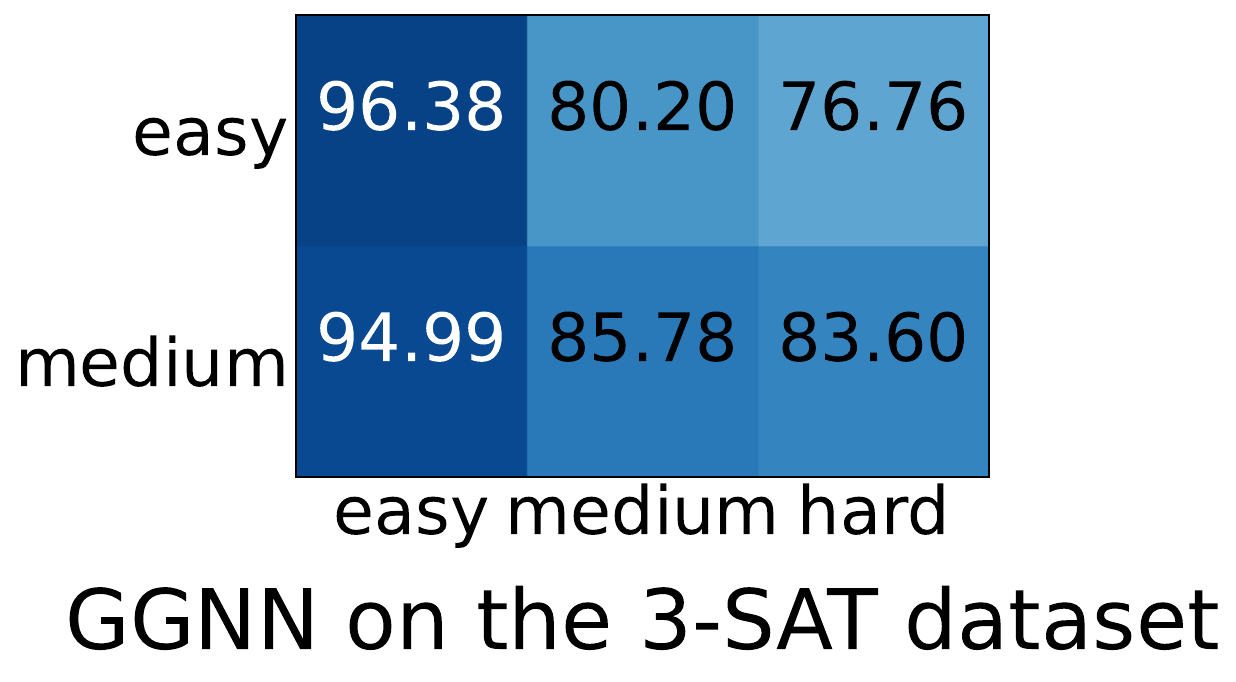}\\
            \end{tabular}
    }
    \end{center}
    \caption{Classification accuracy of satisfiability across different difficulty levels. The x-axis denotes testing datasets and the y-axis denotes training datasets.}
    \label{fig:sat-diff}
    
\end{figure*}

\paragraph{Evaluation across different distributions.}
To assess the generalization ability of GNN models, we evaluate the performance of NeuroSAT (on LCG*) and GGNN (on VCG*) across different datasets and difficulty levels. As shown in Figure~\ref{fig:sat-dataset} and Figure~\ref{fig:sat-diff}, NeuroSAT and GGNN struggle to generalize effectively to datasets distinct from their training data in most cases. However, when trained on the SR dataset, they exhibit better generalization performance across different datasets. Furthermore, while both GNN models demonstrate limited generalization to larger formulas beyond their training data, they perform relatively better on smaller instances. These observations suggest that the generalization performance of GNN models for satisfiability prediction is influenced by the distinct nature and complexity of its training data. Training on more challenging instances could potentially enhance their generalization ability.

Due to the limited space, Figure~\ref{fig:sat-diff} exclusively displays the performance of NeuroSAT and GGNN on the SR and 3-SAT datasets. Comprehensive results on the other five datasets, as well as the experimental results on different massage passing iterations, are provided in Appendix~\ref{app:ben_sat}.

\subsection{Satisfying Assignment Prediction}
\paragraph{Evaluation with different training losses.}
Table~\ref{tab:assignment_general} presents the results of NeuroSAT (on LCG*) and GGNN (on VCG*) across three different training objectives. The results of other GNN models are listed in Table~\ref{tab:assignment_general_app} in Appendix~\ref{app:ben_ass}. Interestingly, the unsupervised training methods outperform the supervised learning approach across the majority of datasets. We hypothesize that this is due to the presence of multiple satisfying assignments in most satisfiable instances. Supervised training tends to bias GNN models towards learning a specific satisfying solution, thereby neglecting the exploration of other feasible ones. This bias may compromise the models' ability to generalize effectively. Such limitations become increasingly apparent when the space of satisfying solutions is much larger, as seen in the medium CA and PS datasets. Additionally, it is noteworthy that employing UNS$_1$ as the loss function can result in instability during the training of some GNN models, leading to a failure to converge in some cases. Conversely, using UNS$_2$ loss demonstrates strong and stable performance across all datasets.

\begin{table}[ht]
  \caption{Solving accuracy on identical distribution with different training losses. SUP denotes the supervised loss, UNS$_{1}$ and UNS$_{2}$ correspond to the unsupervised losses defined in Equation~\ref{eq:unsup} and Equation~\ref{eq:unsupv2}, respectively. The symbol ``-'' indicates that some seeds failed during training. Note that only satisfiable instances are evaluated in this experiment.}
  \label{tab:assignment_general}
  \centering
  \resizebox{\textwidth}{!}{
  \begin{tabular}{lllcccccccccccccc}
    \toprule[1pt]
    \multirow{2}{*}{\bf Graph} & \multirow{2}{*}{\bf Method} & \multirow{2}{*}{\bf Loss} & \multicolumn{7}{c}{\bf Easy Datasets} & \multicolumn{7}{c}{\bf Medium Datasets} \\
    \cmidrule(lr){4-10}
    \cmidrule(lr){11-17}
     & & & \bf SR & \bf 3-SAT & \bf CA & \bf PS & \bf $k$-Clique & \bf $k$-Domset & \bf $k$-Vercov & \bf SR & \bf 3-SAT & \bf CA & \bf PS & \bf $k$-Clique & \bf $k$-Domset & \bf $k$-Vercov \\
    \midrule
    \multirow{3}{*}{LCG*} & \multirow{3}{*}{NeuroSAT} & SUP & \textbf{88.47} & 78.39 & 0.27 & 39.18 & 66.30 & 69.61 & 85.15 & 34.97 & 20.07 & 0.00 & 3.64 & \textbf{56.61} & 52.09 & 74.77  \\
    & & UNS$_{1}$ & 82.30 & 80.23 & 82.17 & \textbf{89.23} & \textbf{88.34} & 96.74 & 99.36 & 25.00 & 30.40 & 35.45 & 60.28 & 41.45 & 95.06 & 67.44 \\
    & & UNS$_{2}$ & 79.79 & \textbf{80.59} & \textbf{89.34} & 88.79 & 63.43 & \textbf{98.85} & \textbf{99.73} & \textbf{37.25} & \textbf{41.61} & \textbf{70.83} & \textbf{71.03} & - & \textbf{96.18} & \textbf{95.99} \\
    \midrule
    \multirow{3}{*}{VCG*} & \multirow{3}{*}{GGNN} & SUP & \textbf{84.13} & 72.87 & 0.29 & 38.82 & 60.80 & 68.36 & 82.06 & 14.15 & 7.96 & 0.00 & 2.33 & 52.35 & 49.07 & 69.21 \\
    & & UNS$_{1}$ & 76.39 & \textbf{76.55} & 78.13 & 84.44 & 84.60 & 97.49 & - & 16.55 & 22.84 & 28.12 & 44.89 & 54.29 & - & 66.37\\
    & & UNS$_{2}$ & 78.75 & 76.42 & \textbf{84.08} & \textbf{86.29} & \textbf{87.12} & \textbf{98.06} & \textbf{99.34} & \textbf{21.18} & \textbf{25.68} & \textbf{50.66} & \textbf{57.96} & \textbf{68.91} & \textbf{92.26} & \textbf{94.30} \\
    \bottomrule[1pt]
  \end{tabular}
  }
\end{table}

In addition to evaluating the performance of GNN models under various training loss functions, we extend our analysis to explore how these models perform across different data distributions and under various inference algorithms. Furthermore, we assess the robustness of these GNN models when trained on noisy datasets that include unsatisfiable instances in an unsupervised fashion. For detailed results of these evaluations across different GNN baselines, please refer to Appendix~\ref{app:ben_ass}.

\subsection{Unsat-core Variable Prediction}
\paragraph{Evaluation on the same distribution.}
The benchmarking results presented in Table~\ref{tab:core_variable_general} exhibit the superior performance of all GNN models on both easy and medium datasets, with NeuroSAT consistently achieving the best results across most datasets. It is important to note that the primary objective of predicting unsat-core variables is not to solve SAT problems directly but to provide valuable guidance for enhancing the backtracking search process. As such, even imperfect predictions - for instance, those with a classification accuracy of 90\% - have been demonstrated to be sufficiently effective in improving the search heuristics employed by modern CDCL-based SAT solvers, as indicated by previous studies~\citep{neurocore,neurocomb}.

\begin{table}[h]
  \caption{Classification accuracy of unsat-core variables on identical distribution. Only unsatisfiable instances are evaluated.}
  \label{tab:core_variable_general}
  \centering
  \resizebox{\textwidth}{!}{
  \begin{tabular}{llcccccccccccccc}
    \toprule[1pt]
    \multirow{2}{*}{\bf Graph} & \multirow{2}{*}{\bf Method} & \multicolumn{7}{c}{\bf Easy Datasets} & \multicolumn{7}{c}{\bf Medium Datasets} \\
    \cmidrule(lr){3-9}
    \cmidrule(lr){10-16}
     & & \bf SR & \bf 3-SAT & \bf CA & \bf PS & \bf $k$-Clique & \bf $k$-Domset & \bf $k$-Vercov & \bf SR & \bf 3-SAT & \bf CA & \bf PS & \bf $k$-Clique & \bf $k$-Domset & \bf $k$-Vercov \\
    \midrule
    \multirow{4}{*}{LCG*} & NeuroSAT & \textbf{90.76} & \textbf{94.43} & \textbf{83.69} & \textbf{86.20} & \textbf{99.93} & 95.80 & \textbf{94.47} & \textbf{90.07} & \textbf{99.65} & \textbf{85.73} & \textbf{88.53} & \textbf{99.97} & \textbf{97.90} & \textbf{99.10} \\
    & GCN & 89.17 & 94.35 & 82.89 & 85.32 & 99.93 & 95.74 & 94.43 & 88.11 & 99.65 & 85.71 & 87.70 & 99.96 & 97.89 & 99.10 \\
    & GGNN & 90.02 & 94.38 & 83.59 & 86.03 & 99.93 & 95.79 & 94.46 & 89.05 & 99.65 & 85.69 & 87.95 & 99.96 & 97.89 & 99.09 \\
    & GIN & 89.29 & 94.33 & 83.71 & 85.97 & 99.93 & \textbf{95.81} & 94.47 & 88.85 & 99.65 & 85.71 & 87.92 & 99.96 & 97.89 & 99.09\\
    \midrule
    \multirow{3}{*}{VCG*} & GCN & 88.57 & 94.34 & 83.17 & 85.27 & 99.93 & \textbf{95.79} & 94.46 & 88.17 & 99.65 & 85.70 & 87.37 & 99.96 & \textbf{97.90} & 99.09 \\
    & GGNN &  \textbf{89.57} & \textbf{94.37} & \textbf{83.50} & \textbf{85.84} & \textbf{99.93} & 95.81 & \textbf{94.49} & 88.84 & \textbf{99.65} & 85.68 & 88.03 & \textbf{99.98} & 97.90 & \textbf{99.10} \\
     & GIN & 89.50 & 94.35 & 83.23 & 85.69 & 99.93 & 95.79 & 94.47 & \textbf{89.51} & 99.65 & \textbf{85.72} & \textbf{88.13} & 99.96 & 97.89 & 99.10 \\
    \bottomrule[1pt]
  \end{tabular}
  }
\end{table}

We also conduct experiments to evaluate the generalization ability of GNN models on unsat-core variable prediction. Please see appendix~\ref{app:ben_unsat} for details.

\section{Advancing Evaluation on \model{}}
To gain deeper insights into how GNNs tackle the SAT problem, we conduct comprehensive comparative analyses between GNN-based SAT solvers and the CDCL and LS heuristics in this section. Since these search heuristics aim to solve a SAT instance directly, our focus only lies on the tasks of (\textbf{T1}) satisfiability prediction and (\textbf{T2}) satisfying assignment prediction (with UNS$_2$ as the training loss). We employ NeuroSAT (on LCG*) and GGNN (on VCG*) as our GNN models and experiment on the SR and 3-SAT datasets. Detailed experimental settings are included in Appendix~\ref{app:adv_exp}.

\subsection{Comparison with the CDCL Heuristic}
\paragraph{Evaluation on the clause-learning augmented instances.}
CDCL-based SAT solvers enhance backtracking search with conflict analysis and clause learning, enabling efficient exploration of the search space by iteratively adding ``learned clauses'' to avoid similar conflicts in future searches~\citep{grasp}. To assess whether GNN-based SAT solvers can learn and benefit from the backtracking search (with CDCL) heuristic, we augment the original formulas in the datasets with learned clauses and evaluate GNN models on these clause-augmented instances. 

Table~\ref{tab:augmented_exp} shows the testing results on augmented SAT datasets. Notably, training on the augmented instances leads to significant improvements in both satisfiability prediction and satisfying assignment prediction. These improvements can be attributed to the presence of ``learned clauses'' that effectively modify the structure of the original formulas, thereby facilitating GNNs to solve with relative ease. However, despite the augmented instances being easily solvable using the backtracking search within a few search steps, GNN models fail to effectively handle these instances when trained on the original instances. These findings suggest that GNNs may not implicitly learn the CDCL heuristic when trained for satisfiability prediction or satisfying assignment prediction.

\begin{table}[ht]
\parbox{0.49\textwidth}{
  \caption{Results on augmented datasets. Values inside/outside parentheses denote the results of models trained on augmented/original instances.}
  \label{tab:augmented_exp}
  \centering
  \resizebox{0.49\textwidth}{!}{
  \begin{tabular}{llcccc}
    \toprule[1pt]
    \multirow{2}{*}{\bf Task} & \multirow{2}{*}{\bf Method} & \multicolumn{2}{c}{\bf Easy Datasets} & \multicolumn{2}{c}{\bf Medium Datasets} \\
    \cmidrule(lr){3-4}
    \cmidrule(lr){5-6}
     & & \bf SR & \bf 3-SAT & \bf SR & \bf 3-SAT  \\
    \midrule
    \multirow{2}{*}{T1} & NeuroSAT & 100.00 (96.78) & 100.00 (96.06) & 100.00 (84.57) & 96.78 (84.85) \\
    & GGNN & 100.00 (97.66) & 100.00 (95.46) & 100.00 (84.01) & 96.29 (85.80) \\
    \midrule\midrule
    \multirow{2}{*}{T2} & NeuroSAT & 85.05 (83.28) & 83.50 (81.04) & 51.95 (45.51) & 39.00 (16.52) \\
    & GGNN & 85.35 (83.42) & 81.56 (79.99) & 44.18 (40.09) & 34.67 (14.75) \\
    \bottomrule[1pt]
  \end{tabular}
  }
}\hfill
\parbox{0.49\textwidth}{
  \caption{Results using contrastive pretraining. Values in parentheses denote the difference between the results without pretraining.}
  \label{tab:pretrain_exp}
  \centering
  \resizebox{0.49\textwidth}{!}{
  \begin{tabular}{llcccc}
    \toprule[1pt]
    \multirow{2}{*}{\bf Task} & \multirow{2}{*}{\bf Method} & \multicolumn{2}{c}{\bf Easy Datasets} & \multicolumn{2}{c}{\bf Medium Datasets} \\
    \cmidrule(lr){3-4}
    \cmidrule(lr){5-6}
     & & \bf SR & \bf 3-SAT & \bf SR & \bf 3-SAT  \\
    \midrule
    \multirow{2}{*}{T1} & NeuroSAT & 96.68 (+0.68) & 96.23 (-0.10) & 78.31 (+0.29) & 85.02 (+0.12) \\
    & GGNN & 96.46 (-0.29) & 96.45 (+0.20) & 76.34 (-0.78) & 85.17 (+0.06)  \\
    \midrule\midrule
    \multirow{2}{*}{T2} & NeuroSAT & 80.54 (+0.75) & 79.71 (-0.88) & 36.42 (-0.83) & 41.23 (-0.38) \\
    & GGNN & 80.66 (-0.34) & 79.23 (-0.09) & 33.44 (+0.07) & 36.39 (+0.44)  \\
    \bottomrule[1pt]
  \end{tabular}
  }
}
\end{table}

\paragraph{Evaluation with contrastive pretraining.}
Observing that GNN models exhibit superior performance on clause-learning augmented SAT instances, there is potential to improve the performance of GNNs by learning a latent representation of the original formula similar to its augmented counterpart. Motivated by this, we also experiment with a contrastive learning approach (i.e., SimCLR~\citep{simclr}) to pretrain the representation of CNF formulas to be close to their augmented ones~\citep{augment}, trying to explicitly embed the CDCL heuristic in the latent space through representation learning.

The results of contrastive pretraining are presented in Table~\ref{tab:pretrain_exp}. In contrast to the findings in~\citet{augment}, our results show limited performance improvement through contrastive pretraining, indicating that GNN models still encounter difficulties in effectively learning the CDCL heuristic in the latent space. This observation aligns with the conclusions drawn in~\citet{gnnmayfail}, which highlight that static GNNs may fail to exactly replicate the same search operations due to the dynamic changes in the graph structure introduced by the clause learning technique.

\subsection{Comparison with the LS Heuristic}
\paragraph{Evaluation with random initialization.}
LS-based SAT solvers typically begin by randomly initializing an assignment and then iteratively flip variables guided by specific heuristics until reaching a satisfying assignment. To compare the behaviors of GNNs with this solving procedure, we first conduct an evaluation of GNN models with randomized initial embeddings in both training and testing, emulating the initialization of LS SAT solvers.

\begin{wraptable}{r}{0.48\textwidth}
  \centering
  \vspace{-15pt}
  \caption{Results using random initialization. Values in parentheses denote the difference between the results with learned initialization.}
  \label{tab:random_init}
  \resizebox{0.48\textwidth}{!}{
  \begin{tabular}{llcccc}
    \toprule[1pt]
    \multirow{2}{*}{\bf Task} & \multirow{2}{*}{\bf Method} & \multicolumn{2}{c}{\bf Easy Datasets} & \multicolumn{2}{c}{\bf Medium Datasets} \\
    \cmidrule(lr){3-4}
    \cmidrule(lr){5-6}
     & & \bf SR & \bf 3-SAT & \bf SR & \bf 3-SAT  \\
    \midrule
    \multirow{2}{*}{T1} & NeuroSAT & 97.24 (+1.24) & 96.44 (+0.11) & 77.29 (-0.91) & 84.85 (-0.05) \\
    & GGNN & 96.78 (+0.03) & 96.38 (+0.13) & 76.97 (-0.15) & 85.80 (+0.69)  \\
    \midrule\midrule
    \multirow{2}{*}{T2} & NeuroSAT & 79.09 (-0.70) & 80.79 (+0.20) & 37.27 (+0.02) & 40.75 (-0.86) \\
    & GGNN & 80.10 (-0.90) & 79.83 (+0.51) & 32.85 (-0.52) & 36.59 (+0.64)  \\
    \bottomrule[1pt]
  \end{tabular}
  }
  \vspace{-10pt}
\end{wraptable}
The results presented in Table~\ref{tab:random_init} demonstrate that using random initialization has a limited impact on the overall performances of GNN-based SAT solvers. This suggests that GNN models do not aim to learn a fixed latent representation of each formula for satisfiability prediction and satisfying assignment prediction. Instead, they have developed a solving strategy that effectively exploits the inherent graph structure of each SAT instance.

\paragraph{Evaluation on the predicted assignments.}
Under random initialization, we further analyze the solving strategies of GNNs by evaluating their predicted assignments decoded from the latent space. For the task of satisfiability prediction, we employ the 2-clustering decoding algorithm to extract the predicted assignments from the literal embeddings of NeuroSAT at each iteration of message passing. For satisfying assignment prediction, we evaluate both NeuroSAT and GGNN using multiple-prediction decoding. Our evaluation focuses on three key aspects: (a) the number of distinct predicted assignments, (b) the number of flipped variables between two consecutive iterations, and (c) the number of unsatisfiable clauses associated with the predicted assignments.

\begin{figure}[ht]
     \centering
     \begin{subfigure}[b]{0.32\textwidth}
         \centering
         \includegraphics[width=\textwidth]{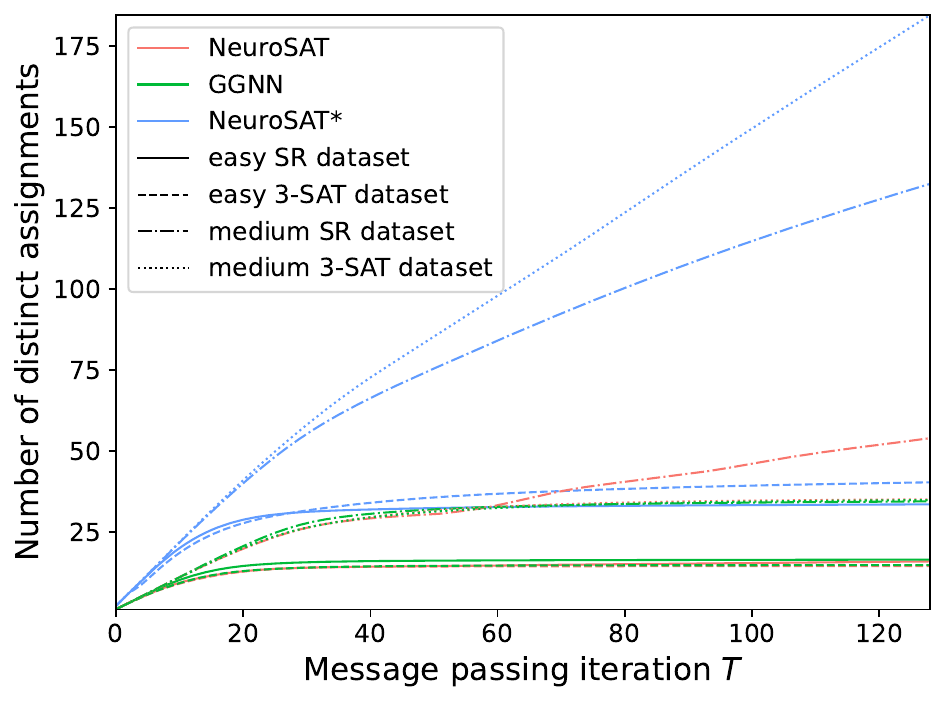}
         \caption{\#Distinct assignments}
         \label{fig:distinct_assignments}
     \end{subfigure}
     \hfill
     \begin{subfigure}[b]{0.32\textwidth}
         \centering
         \includegraphics[width=\textwidth]{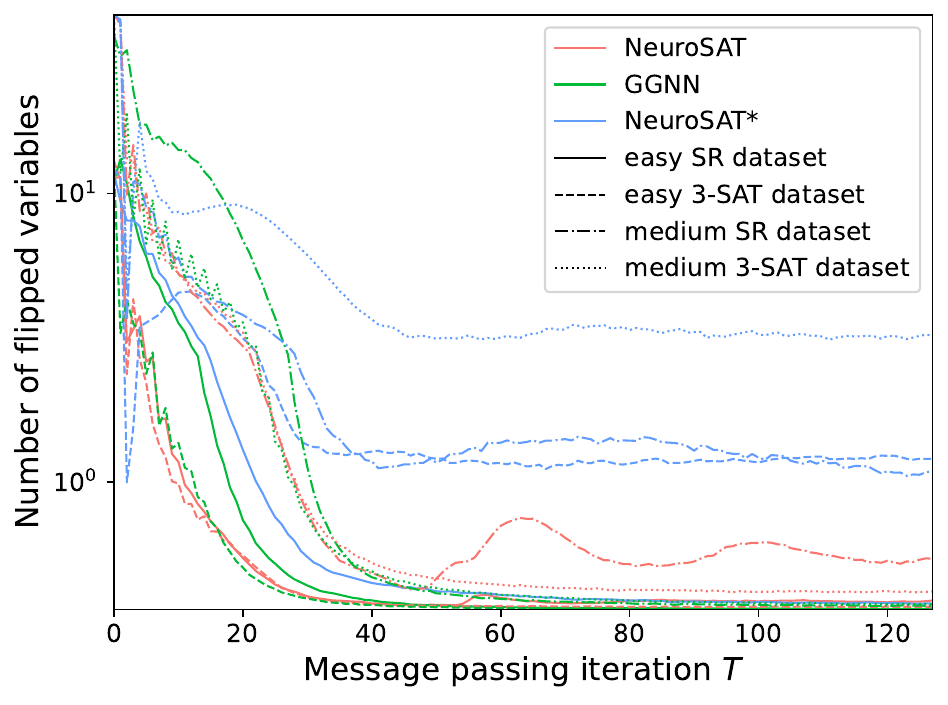}
         \caption{\#Flipped variables}
         \label{fig:flipped_variables}
     \end{subfigure}
     \hfill
     \begin{subfigure}[b]{0.32\textwidth}
         \centering
         \includegraphics[width=\textwidth]{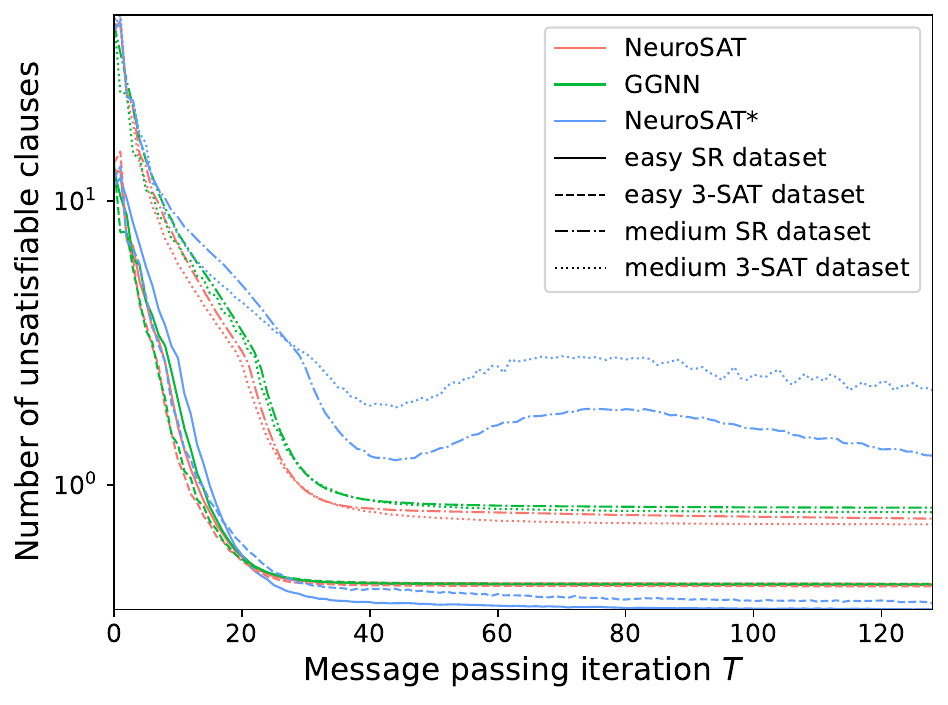}
         \caption{\#Unsatisfible clauses}
         \label{fig:unsat_clauses}
     \end{subfigure}
    \caption{Results on the predicted assignments with the increased message passing iteration $T$. NeuroSAT* refers to the model trained for satisfiability prediction. 
    }
    \label{fig:predicted_assignments}
\end{figure}

As shown in Figure~\ref{fig:predicted_assignments}, all three GNN models initially generate a wide array of assignment predictions by flipping a considerable number of variables, resulting in a notable reduction in the number of unsatisfiable clauses. However, as the iterations progress, the number of flipped variables diminishes substantially, and most GNN models eventually converge towards predicting a specific assignment or making minimal changes to their predictions when there are no or very few unsatisfiable clauses remaining. This trend is reminiscent of the greedy solving strategy adopted by the LS solver GSAT~\citep{gsat}, where changes are made to minimize the number of unsatisfied clauses in the new assignment. However, unlike GSAT's approach of flipping one variable at a time and incorporating random selection to break ties, GNN models simultaneously modify multiple variables and potentially converge to a particular unsatisfied assignment and find it challenging to deviate from such a prediction. It is also noteworthy that despite being trained for satisfiability prediction, NeuroSAT* demonstrates similar behavior to the GNN models trained for assignment prediction. This observation indicates that GNNs also learn to search for a satisfying assignment implicitly in the latent space while performing satisfiability prediction. To provide more insights into the strengths and limitations of GNN-based heuristics, we further conduct experiments to compare GNN-based SAT solvers against state-of-the-art CDCL and LS-based SAT solvers in Appendix~\ref{app:adv_comparisons}.
\section{Discussions}
\subsection{Limitations and Future Work}
\label{app:limitations}
While \model{} represents a significant step in evaluating GNNs for SAT solving, there are still some limitations and potential future directions to consider. \underline{Firstly}, \model{} primarily focuses on evaluating standalone neural SAT solvers, excluding the exploration of neural-guided SAT solvers that integrate GNNs with search-based SAT solvers. It also should be emphasized that the instances included in \model{} are considerably smaller compared to most practical instances found in real-world applications, where GNN models alone are not sufficient for solving such large-scale instances. The efficacy of GNN models in unsat-core prediction shows a promising avenue for combining GNNs with modern SAT solvers, and future research could explore more techniques to effectively leverage these neural-guided SAT solvers to scale up to real-world instances. \underline{Secondly}, \model{} benchmarks general GNN models on the LCG* and VCG* graph representations for SAT solving, but does not consider sophisticated GNN models designed for specific graph constructions in certain domains, such as Circuit SAT problems. Investigating domain-specific GNN models tailored to the characteristics of specific problems could lead to improved performance in specialized instances. \underline{Lastly}, all existing GNN-based SAT solvers in the literature are static GNNs, which have limited learning ability to capture the CDCL heuristic. Exploring dynamic GNN models that can effectively learn the CDCL heuristic is also a potential direction for future research.

\subsection{Conclusion}
\label{sec:conclusion}
In this work, we present \model{}, a benchmark study that comprehensively evaluates GNN models in SAT solving. \model{} offers curated synthetic SAT datasets sourced from various domains and difficulty levels and benchmarks a wide range of GNN-based SAT solvers under diverse settings. Our empirical analysis yields valuable insights into the performances of GNN-based SAT solvers and further provides a deeper understanding of their capabilities and limitations. We hope the proposed \model{} will serve as a solid foundation for GNN-based SAT solving and inspire future research in this exciting field. 

\subsubsection*{Acknowledgments}
This work was supported, in part, by Individual Discovery Grants from the Natural Sciences and Engineering Research Council of Canada, and the Canada CIFAR AI Chair Program.

\bibliography{main}
\bibliographystyle{tmlr}

\clearpage
\appendix
\section{Datasets}
\label{app:dataset}
\paragraph{Generators.} To generate high-quality SAT datasets that do not contain trivial instances, we have employed a rigorous process of selecting appropriate parameters for each CNF generator in \model{}. Table~\ref{tab:generator_settings} provides detailed information about the generators we have used.

\begin{table}[h]
  \caption{Details of the synthetic generators employed in \model{}.}
  \label{tab:generator_settings}
  \centering
  \resizebox{\textwidth}{!}{
  \begin{tabular}{m{4.5em}m{30em}m{17.5em}m{11.5em}}
  \toprule[1pt]
  \multicolumn{1}{l}{\textbf{Dataset}} & \multicolumn{1}{c}{\textbf{Description}} & \multicolumn{1}{c}{\textbf{Parameters}} & \multicolumn{1}{c}{\textbf{Notes}} \\
  \midrule
  SR &
  The SR dataset is composed of pairs of satisfiable and unsatisfiable formulas, with the only difference between each pair being the polarity of a single literal. Given the number of variables $n$, the synthetic generator iteratively samples $k = 1 + \text{Bernoulli}(b) + \text{Geometric}(g)$ variables uniformly at random without replacement and negates each one with independent probability 50\% to build a clause. This procedure continues until the generated formula is unsatisfiable. The satisfiable instance is then constructed by negating the first literal in the last clause of the unsatisfiable one. &
  General: $b = 0.3, g=0.4$, \newline
  Easy dataset: $n \sim \text{Uniform}(10, 40)$, \newline
  Medium dataset: $n \sim \text{Uniform}(40, 200)$, \newline
  Hard dataset: $n \sim \text{Uniform}(200, 400)$ & 
  The sampling parameters are the same as the original paper~\citep{neurosat}.\\
  \midrule
  3-SAT &
  The 3-SAT dataset comprises CNF formulas at the phase transition, where the proportion of generated satisfiable and unsatisfiable formulas is roughly equal. Given the number of variables $n$ and clauses $m$, the synthetic generator iteratively samples three variables (and their polarities) uniformly at random until $m$ clauses are obtained. &
  General: $m = 4.258n +58.26n^{-2/3}$, \newline
  Easy dataset: $n \sim \text{Uniform}(10, 40)$, \newline
  Medium dataset: $n \sim \text{Uniform}(40, 200)$, \newline
  Hard dataset: $n \sim \text{Uniform}(200, 300)$ &
  The parameter $m$ is the same as the paper~\citep{3-sat}\\
  \midrule
  CA & 
  The CA dataset contains SAT instances that are designed to mimic the community structures and modularity features found in real-world industrial instances. Given variable number $n$, clause number $m$, clause size $k$, community number $c$, and modularity $Q$, the synthetic generator iteratively selects $k$ literals in the same community uniformly at random with probability $P = Q + 1/c$ and selects $k$ literals in the distinct community uniformly at random with probability 1 - $P$ to build a clause and repeat for $m$ times to construct a CNF formula. &
  General: $m \sim \text{Uniform}(13n, 15n)$, \newline
  $k \sim \text{Uniform}(4, 5)$, \newline
  $c \sim \text{Uniform}(3, 10)$, \newline
  $Q \sim \text{Uniform}(0.7, 0.9)$ \newline
  Easy dataset: $n \sim \text{Uniform}(10, 40)$, \newline
  Medium dataset: $n \sim \text{Uniform}(40, 200)$, \newline
  Hard dataset: $n \sim \text{Uniform}(200, 400)$ &
  The parameters are selected based on the experiments in the original paper~\citep{ca} and our study to ensure that the generated SAT instances have a balance of satisfiability and unsatisfiability.\\
  \midrule
  PS &
  PS dataset encompasses SAT instances with a power-law distribution in the number of variable occurrences (popularity), and good clustering between them (similarity). Given variable number $n$, clause number $m$, and average clause size $k$, the synthetic generator first assigns random angles $\theta_i, \theta_j \in [0, 2\pi]$ to each variable $i$ and each clause $j$, and then randomly samples variable $i$ in clause $j$ with the probability $P = 1 / (1 + (i^{\beta}j^{\beta'}\theta_{ij}/R)^T)$. Here, $\theta_{ij} = \pi - |\pi - |\theta_i - \theta_j||$ is the angle between variable $i$ and clause $j$. The exponent parameters $\beta$ and $\beta'$ control the power-law distribution of variable occurrences and clause size respectively. The temperature parameter $T$ controls the sharpness of the probability distribution, while $R$ is an approximate normalization constant that ensures the average number of selected edges is $km$. &
  General: $m \sim \text{Uniform}(6n, 8n)$, \newline
  $k \sim \text{Uniform}(4, 5)$, \newline
  $\beta \sim \text{Uniform}(0, 1)$, \newline
  $\beta' = 1$, \newline
  $c \sim \text{Uniform}(3, 10)$, \newline
  $T \sim \text{Uniform}(0.75, 1.5)$ \newline
  Easy dataset: $n \sim \text{Uniform}(10, 40)$, \newline
  Medium dataset: $n \sim \text{Uniform}(40, 200)$, \newline
  Hard dataset: $n \sim \text{Uniform}(200, 300)$ &
  The parameters are selected based on the experiments in the original paper~\citep{ps} and our study to ensure that the generated SAT instances have a balance of satisfiability and unsatisfiability.\\
  \midrule
  $k$-Clique &
  The $k$-Clique dataset includes SAT instances that encode the $k$-Clique problem, which involves determining whether there exists a clique (i.e., a subset of vertices that are all adjacent to each other) with $v$ vertices in a given graph. Given the number of cliques $k$, the synthetic generator produces an Erd\H{o}s-R\'{e}nyi graph with $v$ vertices and a given edge probability $p$ and then transforms the corresponding $k$-Clique problem into a SAT instance. &
  General: $p = \binom{v}{k}^{-1/\binom{v}{2}}$, \newline
  Easy dataset: $v \sim \text{Uniform}(5, 15)$, \newline
  $k \sim \text{Uniform}(3, 4)$, \newline
  Medium dataset: $v \sim \text{Uniform}(15, 20)$, \newline
  $k \sim \text{Uniform}(3, 5)$, \newline
  Hard dataset: $v \sim \text{Uniform}(20, 25)$, \newline
  $k \sim \text{Uniform}(4, 6)$ &
  The parameter $p$ is selected based on the paper~\citep{k-cliques}, making the expected number of $k$-Cliques in the generated graph equals 1.\\
  \midrule
  $k$-Domset &
  The $k$-Domset dataset contains SAT instances that encode the $k$-Dominating Set problem. This problem is to determine whether there exists a dominating set (i.e., a subset of vertices such that every vertex in the graph is either in the subset or adjacent to a vertex in the subset) with at most $k$ vertices in a given graph. Given the domination number $k$, the synthetic generator produces an Erd\H{o}s-R\'{e}nyi graph with $v$ vertices and a given edge probability $p$ and then transforms the corresponding $k$-Dominating Set problem into a SAT instance. &
  General: $p = 1 - \left(1 - \binom{v}{k}^{-1/(v-k)}\right)^{1/k}$, \newline
  Easy dataset: $v \sim \text{Uniform}(5, 15)$, \newline
  $k \sim \text{Uniform}(2, 3)$, \newline
  Medium dataset: $v \sim \text{Uniform}(15, 20)$, \newline
  $k \sim \text{Uniform}(3, 5)$, \newline
  Hard dataset: $v \sim \text{Uniform}(20, 25)$, \newline
  $k \sim \text{Uniform}(4, 6)$ &
  The parameter $p$ is selected based on the paper~\citep{k-domset}, making the expected number of domination set with size $k$ in the generated graph equals 1.\\
  \midrule
  $k$-Vercov &
  The $k$-Vercov dataset consists of SAT instances that encode the $k$-Vertex Cover problem, i.e., check whether there exists a set of $k$ vertices in a graph such that every edge has at least one endpoint in this set. Given the vertex cover number $k$, the synthetic generator produces a complement graph of an Erd\H{o}s-R\'{e}nyi graph with $v$ vertices and a given edge probability $p$ and then converts the corresponding $k$-Vertex Cover problem into a SAT instance. &
  General: $p = \binom{v}{k}^{-1/\binom{v}{2}}$, \newline
  Easy dataset: $v \sim \text{Uniform}(5, 15)$, \newline
  $k \sim \text{Uniform}(3, 5)$, \newline
  Medium dataset: $v \sim \text{Uniform}(10, 20)$, \newline
  $k \sim \text{Uniform}(6, 8)$, \newline
  Hard dataset: $v \sim \text{Uniform}(15, 25)$, \newline
  $k \sim \text{Uniform}(9, 10)$ &
  The generation process and the parameter are selected based on the relationship between $k$-Vertex Cover and $k$-Clique problems, making the size of the minimum vertex cover in the generated graph around $k$. \\
  \bottomrule[1pt]
  \end{tabular}
  }
\end{table}

\paragraph{Statistics.}
To provide a comprehensive understanding of our generated datasets, we compute several characteristics across three difficulty levels. These statistics include the average number of variables and clauses, as well as graph measures such as average clustering coefficient (in VIG) and modularity (in VIG, VCG, and LCG). The dataset statistics are summarized in Table~\ref{tab:dataset_statistics}.

\begin{table}[ht]
  \caption{Dataset statistics across difficulty levels in \model{}.}
  \label{tab:dataset_statistics}
  \centering
  \small
  \resizebox{\textwidth}{!}{
  \begin{tabular}{lcccccccccccccccccc}
    \toprule[1pt]
    \multirow{2}{*}{\bf Dataset} & \multicolumn{6}{c}{\bf Easy Difficulty} & \multicolumn{6}{c}{\bf Medium Difficulty} & \multicolumn{6}{c}{\bf Hard Difficulty} \\
    \cmidrule(lr){2-7}
    \cmidrule(lr){8-13}
    \cmidrule(lr){14-19}
    & \bf \#Variables & \bf \#Clauses & \bf C.C.(VIG) & \bf Mod.(VIG) & \bf Mod.(VCG) & \bf Mod.(LCG) & \bf \#Variables & \bf \#Clauses & \bf C.C.(VIG) & \bf Mod.(VIG) & \bf Mod.(VCG) & \bf Mod.(LCG) & \bf \#Variables & \bf \#Clauses & \bf C.C.(VIG) & \bf Mod.(VIG) & \bf Mod.(VCG) & \bf Mod.(LCG) \\
    \midrule
    SR & 25.00 & 148.35 & 0.98 & 0.00 & 0.25 & 0.33 & 118.36 & 646.54 & 0.62 & 0.06 & 0.31 & 0.37 & 299.64 & 1613.86 & 0.32 & 0.09 & 0.32 & 0.37 \\
    \midrule
    3-SAT & 25.05 & 113.69 & 0.72 & 0.06 & 0.36 & 0.46 & 120.00 & 513.14 & 0.27 & 0.16 & 0.43 & 0.51 & 250.44 & 1067.34 & 0.14 & 0.17 & 0.45 & 0.52 \\
    \midrule
    CA & 31.66 & 303.48 & 0.65 & 0.19 & 0.73 & 0.73 & 120.27 & 1661.07 & 0.54 & 0.38 & 0.80 & 0.80 & 299.68 & 4195.50 & 0.59 & 0.57 & 0.80 & 0.80 \\
    \midrule
    PS & 25.41 & 176.68 & 0.98 & 0.00 & 0.27 & 0.32 & 118.75 & 822.78 & 0.86 & 0.05 & 0.35 & 0.37 & 249.61 & 1728.34 & 0.77 & 0.08 & 0.38 & 0.28 \\
    \midrule
    $k$-Clique & 34.85 & 592.89 & 0.90 & 0.03 & 0.45 & 0.49 & 69.56 & 2220.05 & 0.91 & 0.03 & 0.48 & 0.49 & 112.87 & 5543.26 & 0.88 & 0.04 & 0.49 & 0.50 \\
    \midrule
    $k$-Domset & 41.90 & 369.40 & 0.70 & 0.26 & 0.47 & 0.53 & 90.64 & 1736.22 & 0.70 & 0.21 & 0.49 & 0.51 & 137.31 & 4032.48 & 0.70 & 0.20 & 0.49 & 0.51 \\
    \midrule
    $k$-Vercov & 45.41 & 484.28 & 0.66 & 0.16 & 0.48 & 0.53 & 107.40 & 2634.14 & 0.69 & 0.16 & 0.49 & 0.51 & 190.24 & 8190.94 & 0.69 & 0.16 & 0.50 & 0.51 \\
    \bottomrule[1pt]
    \end{tabular}
    }
\end{table}

\section{GNN Models}
\label{app:gnn}
\paragraph{Message-passing schemes on VCG*.}
Recall that VCG* incorporates two distinct edge types, \model{} employs different functions to execute heterogeneous message-passing in each direction of each edge type. Formally, we define a $d$-dimensional embedding for each variable and clause node, denoted by $h_l$ and $h_c$, respectively. These embeddings are initialized to two learnable vectors $h_v^0$ and $h_c^0$, depending on the node type. At the $k$-th iteration of message passing, these hidden representations are updated as follows:
\begin{equation}
\label{eqn:gnn_vcg}
\begin{aligned}
h_c^{(k)} &= \text{UPD}\left(\underset{v \in c^{+}}{\text{AGG}}\left(\left\{\text{MLP}_v^{+}\left(h_v^{(k-1)}\right)\right\}\right), \underset{v \in c^{-}}{\text{AGG}}\left(\left\{\text{MLP}_v^{-}\left(h_v^{(k-1)}\right)\right\}\right), h_c^{(k-1)}\right), \\
h_v^{(k)} &= \text{UPD}\left(\underset{c \in v^{+}}{\text{AGG}}\left(\left\{\text{MLP}_c^{+}\left(h_c^{(k-1)}\right)\right\}\right), \underset{c \in v^{-}}{\text{AGG}}\left(\left\{\text{MLP}_c^{-}\left(h_c^{(k-1)}\right)\right\}\right), h_v^{(k-1)}\right),
\end{aligned}
\end{equation}
where $c^{+}$ and $c^{-}$ denote the sets of variable nodes that occur in the clause $c$ with positive and negative polarity, respectively. Similarly, $v^{+}$ and $v^{-}$ denote the sets of clause nodes where variable $v$ occurs in positive and negative form. $\text{MLP}_v^{+}$, $\text{MLP}_v^{-}$, $\text{MLP}_c^{+}$, and $\text{MLP}_c^{-}$ are four MLPs. $\text{UPD}(\cdot)$ is the update function, and $\text{AGG}(\cdot)$ is the aggregation function.

\paragraph{GNN baselines.}
Table~\ref{tab:gnn_baselines} summarizes the message-passing algorithms of the GNN models used in \model{}. We adopt heterogeneous versions of GCN~\citep{gcn}, GGNN~\citep{ggnn}, and GIN~\citep{gin} on both LCG* and VCG*, while maintaining the original NeuroSAT~\citep{neurosat} only on LCG*.
\begin{table}[ht]
  \caption{Supported GNN models in \model{}.}
  \label{tab:gnn_baselines}
  \centering
  \resizebox{\textwidth}{!}{
  \begin{tabular}{m{3em}m{4em}p{36em}m{20em}}
    \toprule[1pt]
    \multicolumn{1}{l}{\textbf{Graph}} & \multicolumn{1}{l}{\textbf{Method}} & \multicolumn{1}{c}{\textbf{Message-passing Algorithm}} & \multicolumn{1}{c}{\textbf{Notes}}\\
    \midrule
    \multirow{4}{*}[-8.2em]{LCG*} & NeuroSAT & 
    \makecell[l]{$h_c^{(k)}, s_c^{(k)} = \text{LayerNormLSTM}_1\left(\underset{l \in \mathcal{N}(c)}{\sum}\text{MLP}_l\left(h_l^{(k-1)}\right), \left(h_c^{(k-1)}, s_c^{(k-1)}\right)\right)$, \\
    $h_l^{(k)}, s_l^{(k)} = \text{LayerNormLSTM}_2\left(\left[\underset{c \in \mathcal{N}(l)}{\sum}\text{MLP}_c\left(h_c^{(k-1)}\right), h_{\neg l}^{(k-1)}\right], \left(h_l^{(k-1)}, s_l^{(k-1)}\right)\right)$} &
    $s_c, s_l$ are the hidden states which are initialized to zero vectors. \\
    \cmidrule{2-4}
    & GCN &
    \makecell[l]{$h_c^{(k)} = \text{Linear}_1\left(\left[\underset{l \in \mathcal{N}(c)}{\sum} \frac{\text{MLP}_l\left(h_l^{(k-1)}\right)}{\sqrt{d_l d_c}}, h_c^{(k-1)}\right]\right)$, \\
    $h_l^{(k)} = \text{Linear}_2\left(\left[\underset{c \in \mathcal{N}(l)}{\sum} \frac{\text{MLP}_c\left(h_c^{(k-1)}\right)}{\sqrt{d_c d_l}}, h_{\neg l}^{(k-1)}, h_l^{(k-1)}\right]\right)$} &
    $d_c$, $d_l$ are the degrees of clause node $c$ and literal node $l$ in LCG respectively.\\
    \cmidrule{2-4}
    & GGNN &
    \makecell[l]{$h_c^{(k)} = \text{GRU}_1\left(\underset{l \in \mathcal{N}(c)}{\sum}\left(\left\{\text{MLP}_l\left(h_l^{(k-1)}\right)\right\}\right), h_c^{(k-1)}\right)$, \\
    $h_l^{(k)} = \text{GRU}_2\left(\left[\underset{c \in \mathcal{N}(l)}{\sum}\text{MLP}_c\left(h_c^{(k-1)}\right), h_{\neg l}^{(k-1)}\right], h_l^{(k-1)}\right)$} &
    \\
    \cmidrule{2-4}
    & GIN & 
    \makecell[l]{$h_c^{(k)} = \text{MLP}_1\left(\left[\underset{l \in \mathcal{N}(c)}{\sum}\left(\left\{\text{MLP}_l\left(h_l^{(k-1)}\right)\right\}\right), h_c^{(k-1)}\right]\right)$, \\
    $h_l^{(k)} = \text{MLP}_2\left(\left[\underset{c \in \mathcal{N}(l)}{\sum}\text{MLP}_c\left(h_c^{(k-1)}\right), h_{\neg l}^{(k-1)}, h_l^{(k-1)}\right]\right)$} &
    \\
    \midrule
    \multirow{3}{*}[-5.4em]{VCG*} & GCN & 
    \makecell[l]{$h_c^{(k)} = \text{Linear}_1\left(\left[\underset{v \in c^{+}}{\sum}\frac{\text{MLP}_v^{+}\left(h_v^{(k-1)}\right)}{\sqrt{d_v d_c}}, \underset{v \in c^{-}}{\sum}\frac{\text{MLP}_v^{-}\left(h_v^{(k-1)}\right)}{\sqrt{d_v d_c}}, h_c^{(k-1)}\right]\right)$, \\
    $h_v^{(k)} = \text{Linear}_2\left(\left[\underset{c \in v^{+}}{\sum}\frac{\text{MLP}_c^{+}\left(h_c^{(k-1)}\right)}{\sqrt{d_c d_v}}, \underset{c \in v^{-}}{\sum}\frac{\text{MLP}_c^{-}\left(h_c^{(k-1)}\right)}{\sqrt{d_c d_v}}, h_v^{(k-1)}\right]\right)$} &
    $d_c$, $d_v$ are the degrees of clause node $c$ and variable node $v$ in VCG respectively.\\
    \cmidrule{2-4}
    & GGNN & 
    \makecell[l]{$h_c^{(k)} = \text{GRU}_1\left(\left[\underset{v \in c^{+}}{\sum}\text{MLP}_v^{+}\left(h_v^{(k-1)}\right), \underset{v \in c^{-}}{\sum}\text{MLP}_v^{-}\left(h_v^{(k-1)}\right)\right], h_c^{(k-1)}\right)$, \\
    $h_v^{(k)} = \text{GRU}_2\left(\left[\underset{c \in v^{+}}{\sum}\text{MLP}_c^{+}\left(h_c^{(k-1)}\right), \underset{c \in v^{-}}{\sum}\text{MLP}_c^{-}\left(h_c^{(k-1)}\right)\right], h_v^{(k-1)}\right)$} &
    \\
    \cmidrule{2-4}
    & GIN & 
    \makecell[l]{$h_c^{(k)} = \text{MLP}_1\left(\left[\underset{v \in c^{+}}{\sum}\text{MLP}_v^{+}\left(h_v^{(k-1)}\right), \underset{v \in c^{-}}{\sum}\text{MLP}_v^{-}\left(h_v^{(k-1)}\right), h_c^{(k-1)}\right]\right)$, \\
    $h_v^{(k)} = \text{MLP}_2\left(\left[\underset{c \in v^{+}}{\sum}\text{MLP}_c^{+}\left(h_c^{(k-1)}\right), \underset{c \in v^{-}}{\sum}\text{MLP}_c^{-}\left(h_c^{(k-1)}\right), h_v^{(k-1)}\right]\right)$} &
    \\
    \bottomrule[1pt]
  \end{tabular}
  }
\end{table}

\section{Benchmarking Evaluation}
\subsection{Implementation Details}
\label{app:ben_imp}
In \model{}, we provide the ground truth of satisfiability and satisfying assignments by calling the state-of-the-art modern SAT solver CaDiCaL~\citep{cadical} and generate the truth labels for unsat-core variables by invoking the proof checker DRAT-trim~\citep{drat}.
All neural networks in our study are implemented using PyTorch~\citep{torch} and PyTorch Geometric~\citep{pyg}. For all GNN models, we set the feature dimension $d$ to 128 and the number of message passing iterations $T$ to 32. The MLPs in the models consist of two hidden layers with the ReLU~\citep{relu} activation function. To select the optimal hyperparameters for each GNN baseline, we conduct a grid search over several settings. Specifically, we explore different learning rates from $\{10^{-3}, 5\times 10^{-4}, 10^{-4}, 5\times 10^{-5}, 10^{-5}\}$, training epochs from $\{50, 100, 200\}$, weight decay values from $\{10^{-6},10^{-7},10^{-8},10^{-9},10^{-10}\}$, and gradient clipping norms from $\{0.1, 0.5, 1\}$. We employ Adam~\citep{adam} as the optimizer and set the batch size to 128, 64, or 32 to fit within the maximum GPU memory (48G). For the parameters $\tau$ and $\kappa$ of the unsupervised loss in Equation~\ref{eq:smooth_operators} and Equation~\ref{eq:unsup}, we try the default settings ($\tau=t^{-0.4}$ and $\kappa=10$, where $t$ is the global step during training) as the original paper~\citep{circuitsat} as well as other values ($\tau \in \{0.05, 0.1, 0.2, 0.5\}$, $\kappa \in \{1, 2, 5\}$) and empirically find $\tau = 0.1, \kappa=1$ yield the best results. Furthermore, it is important to note that we use three different random seeds to benchmark the performance of different GNN models and assess the generalization ability of NeuroSAT and GGNN using one seed for simplicity.

\subsection{Satiafiability Prediction}
\label{app:ben_sat}
\begin{figure}[h]
    \centering
    \includegraphics[width=\textwidth]{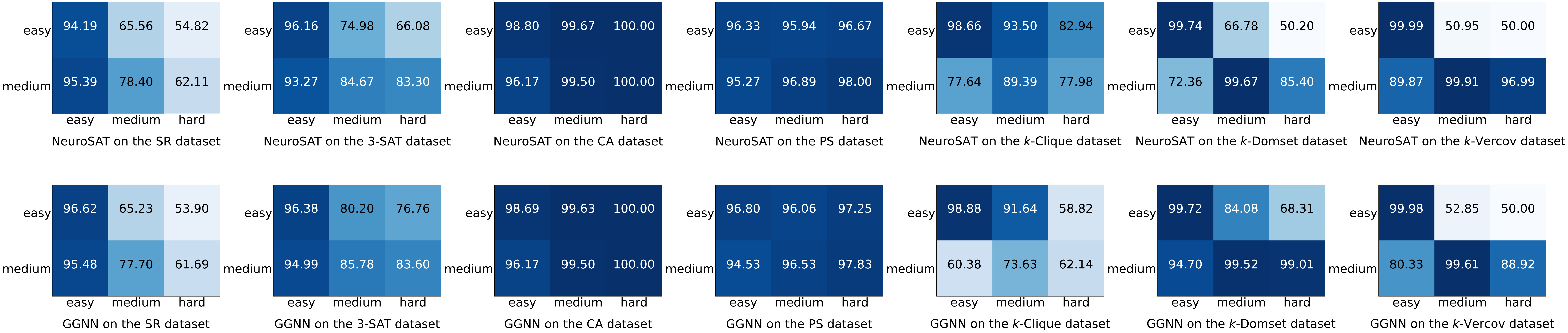}
    \caption{Classification accuracy of satisfiability across different difficulty levels. The x-axis denotes testing datasets and the y-axis denotes training datasets.}
    \label{fig:sat_difficulty_transfer}
\end{figure}

\paragraph{Evaluation across different difficulty levels.}
The complete results of NeuroSAT and GGNN across different difficulty levels are presented in Figure~\ref{fig:sat_difficulty_transfer}. Consistent with the findings on the SR and 3-SAT datasets, both GNN models exhibit limited generalization ability to larger instances beyond their training data, while displaying relatively better performance on smaller instances. This observation suggests that training these models on more challenging instances could potentially enhance their generalization ability and improve their performance on larger instances.

\begin{figure}[h]
    \centering
    \includegraphics[width=\textwidth]{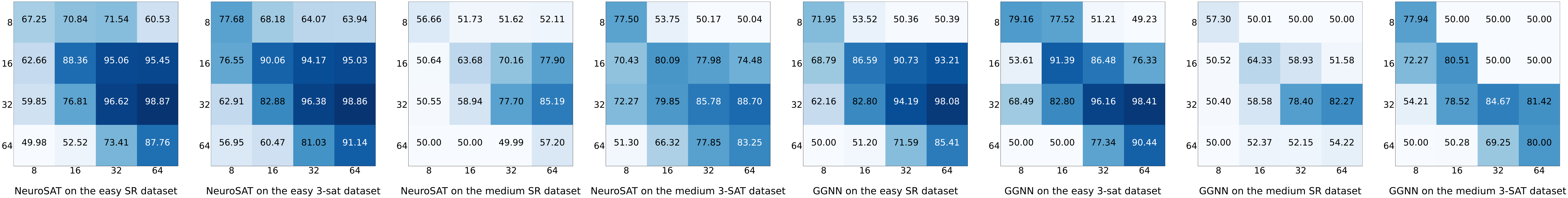}
    \caption{Classification accuracy of satisfiability across different message passing iterations $T$. The x-axis denotes testing iterations and the y-axis denotes training iterations.}
    \label{fig:sat_ite_transfer}
\end{figure}

\paragraph{Evaluation with different message passing iterations.}
To investigate the impact of message-passing iterations on the performance of GNN models during training and testing, we conducted experiments with varying iteration values. Figure~\ref{fig:sat_ite_transfer} presents the results of NeuroSAT and GGNN trained and evaluated with different message passing iterations. Remarkably, using a training iteration value of 32 consistently yielded the best performance for both models. Conversely, employing too small or too large iteration values during training resulted in decreased performance. Furthermore, the models trained with 32 iterations also demonstrated good generalization ability to testing iterations 16 and 64. These findings emphasize the critical importance of selecting an appropriate message-passing iteration to ensure optimal learning and reasoning within GNN models.

\subsection{Satisfying Assignment Prediction}
\label{app:ben_ass}

\paragraph{Evaluation with different training losses.} Table~\ref{tab:assignment_general_app} presents the complete results of each GNN baseline across three different training objectives. Like the results of NeuroSAT and GGNN, all other GNN models with unsupervised training outperform their supervised training counterparts.

\begin{table}[ht]
  \caption{Solving accuracy on identical distribution with different training losses. The top and bottom 7 rows represent the results for easy and medium datasets, respectively. SUP denotes the supervised loss, UNS$_{1}$ and UNS$_{2}$ correspond to the unsupervised losses defined in Equation~\ref{eq:unsup} and Equation~\ref{eq:unsupv2}, respectively. The symbol ``-'' indicates that some seeds failed during training. Note that only satisfiable instances are evaluated in this experiment.}
  \label{tab:assignment_general_app}
  \centering
  \resizebox{\textwidth}{!}{
  \begin{tabular}{llcccccccccccccccccccccc}
    \toprule[1pt]
    \multirow{2}{*}{\bf Graph} & \multirow{2}{*}{\bf Method} & \multicolumn{3}{c}{\bf SR} & \multicolumn{3}{c}{\bf 3-SAT} & \multicolumn{3}{c}{\bf CA} & \multicolumn{3}{c}{\bf PS} & \multicolumn{3}{c}{\bf $k$-Clique} & \multicolumn{3}{c}{\bf $k$-Domset} & \multicolumn{3}{c}{\bf $k$-Vercov} \\
    \cmidrule(lr){3-5}
    \cmidrule(lr){6-8}
    \cmidrule(lr){9-11}
    \cmidrule(lr){12-14}
    \cmidrule(lr){15-17}
    \cmidrule(lr){18-20}
    \cmidrule(lr){21-23}
     & & \bf SUP & \bf UNS$_{1}$ & \bf UNS$_{2}$ & \bf SUP & \bf UNS$_{1}$ & \bf UNS$_{2}$ & \bf SUP & \bf UNS$_{1}$ & \bf UNS$_{2}$ & \bf SUP & \bf UNS$_{1}$ & \bf UNS$_{2}$ & \bf SUP & \bf UNS$_{1}$ & \bf UNS$_{2}$ & \bf SUP & \bf UNS$_{1}$ & \bf UNS$_{2}$ & \bf SUP & \bf UNS$_{1}$ & \bf UNS$_{2}$\\
    \midrule
    \multirow{4}{*}{LCG*} & NeuroSAT & \textbf{88.47} & 82.30 & 79.79 & 78.39 & 80.23 & 80.59 & 0.27 & 82.17 & \textbf{89.34} & 39.18 & \textbf{89.23} & 88.79 & 66.30 & \textbf{88.34} & 63.43 & 69.61 & 96.74 & \textbf{98.85} & 85.15 & 99.36 & \textbf{99.73} \\
    & GCN & 83.74 & 73.09 & 77.02 & 70.34 & 74.79 & 75.31 & 0.17 & 75.30 & 82.41 & 39.66 & 82.75 & 84.89 & 63.85 & 82.60 & 86.17 & 59.29 & 97.50 & 97.55 & 76.83 & 99.16 & 99.28 \\
    & GGNN & 84.13 & 76.39 & 78.75 & 72.87 & 76.55 & 76.42 & 0.29 & 78.13 & 84.08 & 38.82 & 84.44 & 86.29 & 60.80 & 84.60 & 87.12 & 68.36 & 97.49 & 98.06 & 82.06 & - & 99.34 \\
    & GIN & 83.81 & 81.45 & 80.39 & 73.99 & 78.47 & 76.24 & 0.20 & 78.44 & 85.15 & 39.13 & 85.31 & 85.43 & 56.85 & 84.48 & 85.11 & 68.93 & 96.99 & 97.43 & 81.49 & 99.28 & 99.38\\
    \midrule
    \multirow{3}{*}{VCG*} & GCN & 83.38 & 84.19 & 78.00 & 76.60 & 84.42 & 79.23 & 14.98 & 76.64 & 83.79 & 51.48 & 85.88 & 83.06 & 56.27 & 85.28 & 86.91 & 66.32 & 97.62 & 96.74 & 78.67 & - & 93.51 \\
    & GGNN & 86.30 & 87.16 & 81.00 & 77.96 & \textbf{88.97} & 79.32 & 15.11 & 76.32 & 83.12 & 47.67 & 86.85 & 87.17 & 66.86 & 86.31 & 87.48 & 66.42 & - & 98.42 & 82.61 & - & 99.52 \\
     & GIN & 84.61 & 89.56 & 83.27 & 79.23 & 87.65 & 81.72 & 17.81 & 83.28 & 86.03 & 48.92 & 91.21 & 85.65 & 66.07 & 86.12 & 88.09 & 67.67 & - & - & 81.01 & 99.38 & 99.41 \\
    \midrule\midrule
    \multirow{4}{*}{LCG*} & NeuroSAT & 34.97 & 25.00 & \textbf{37.25} & 20.07 & 30.40 & \textbf{41.61} & 0.00 & 35.45 & \textbf{70.83} & 3.64 & 60.28 & \textbf{71.03} & 56.61 & 41.45 & - & 52.09 & 95.06 & \textbf{96.18} & 74.77 & 67.44 & 95.99 \\
    & GCN & 13.19 & 13.76 & 19.21 & 8.87 & 20.50 & 24.58 & 0.00 & 30.20 & 54.04 & 1.45 & 45.16 & 56.29 & 55.36 & 61.82 & 66.33 & 43.50 & 92.86 & 94.89 & 67.83 & - & 93.84 \\
    & GGNN & 14.15 & 16.55 & 21.18 & 7.96 & 22.84 & 25.68 & 0.00 & 28.12 & 50.66 & 2.33 & 44.89 & 57.96 & 52.35 & 54.29 & \textbf{68.91} & 49.07 & - & 92.26 & 69.21 & 66.37 & 94.30 \\
    & GIN & 15.36 & 18.60 & 22.17 & 9.66 & 21.38 & 24.93 & 0.00 & 35.76 & 57.81 & 2.02 & 43.43 & 57.62 & 53.07 & 44.60 & 66.32 & 44.39 & 93.3 & 93.82 & 70.59 & 55.59 & 95.69\\
    \midrule
    \multirow{3}{*}{VCG*} & GCN & 20.59 & 9.21 & 22.44 & 12.48 & 17.00 & 29.53 & 0.44 & 39.04 & 48.99 & 2.29 & 35.99 & 55.46 & 46.09 & 25.90 & 68.62 & 46.96 & - & 92.68 & 69.15 & - & 96.46 \\
    & GGNN & 28.04 & 27.72 & 33.37 & 16.46 & 29.65 & 35.95 & 0.56 & 48.13 & 49.93 & 3.12 & 51.73 & 65.11 & 44.26 & 48.92 & 56.43 & 51.01 & - & - & 71.97 & - & 95.23 \\
     & GIN & 26.73 & 26.48 & 31.97 & 14.64 & 26.86 & 35.81 & 0.64 & 44.06 & 63.84 & 3.38 & 58.03 & 64.66 & 55.47 & 56.97 & 67.78 & 46.98 & - & 95.28 & 69.40 & - & \textbf{96.96} \\
    \bottomrule[1pt]
  \end{tabular}
    }
\end{table}

\begin{figure}[h]
    \centering
    \includegraphics[width=\textwidth]{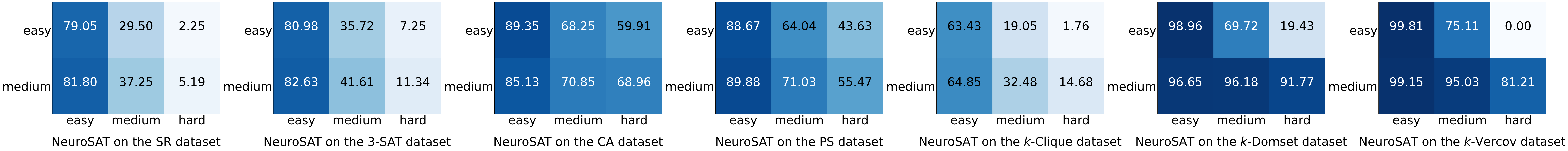}
    \caption{Solving accuracy of NeuroSAT across different difficulty levels (with UNS$_2$ as the training loss). The x-axis denotes testing datasets and the y-axis denotes training datasets.}
    \label{fig:assignment_difficulty_transfer}
\end{figure}

\paragraph{Evaluation across different difficulty levels.}
The performance of NeuroSAT across different difficulty levels is shown in Figure~\ref{fig:assignment_difficulty_transfer}. Notably, training on medium datasets yields superior generalization performance compared to training on easy datasets. This suggests that training on more challenging SAT instances with larger sizes can enhance the model's ability to generalize to a wider range of problem complexities.

\begin{wrapfigure}{r}{0.48\textwidth}
\centering
\includegraphics[width=0.48\textwidth]{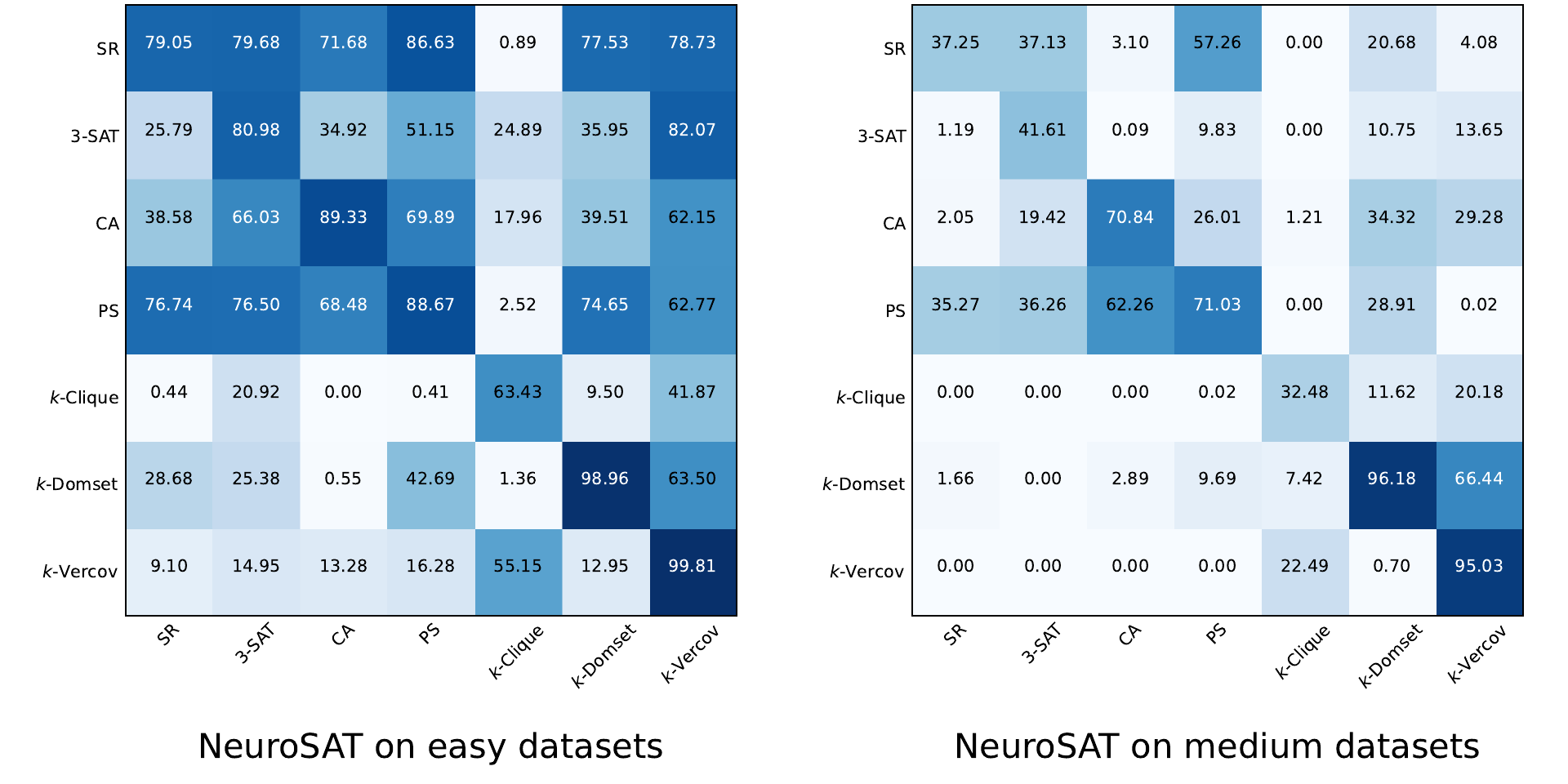}
\caption{Solving accuracy of NeuroSAT across different datasets (with UNS$_2$ as the training loss). The x-axis denotes testing datasets and the y-axis denotes training datasets.}
\label{fig:assignment_dataset_transfer}
\vspace{-5pt}
\end{wrapfigure}
\paragraph{Evaluation with different datasets.}
Figure~\ref{fig:assignment_dataset_transfer} illustrates the performance of NeuroSAT across different datasets. For easy datasets, we observe that NeuroSAT demonstrates a strong generalization ability to other datasets when trained on the SR, 3-SAT, CA, and PS datasets. However, when trained on the $k$-Clique, $k$-Domset, and $k$-Vercov datasets, which involve specific graph structures inherent to their combinatorial problems, NeuroSAT struggles to generalize effectively. This observation indicates that the GNN model may overfit to leverage specific graph features associated with these combinatorial datasets, without developing a generalized solving strategy that can be applied to other problem domains for satisfying assignment prediction. For medium datasets, NeuroSAT also faces challenges in generalization, as its performance is relatively limited. This can be attributed to the difficulty of these datasets, where finding satisfying assignments is much harder than easy datasets.

\paragraph{Evaluation with different inference algorithms.}
\begin{wrapfigure}{r}{0.48\textwidth}
\vspace{-17.5pt}
\begin{center}
\includegraphics[width=0.48\textwidth]{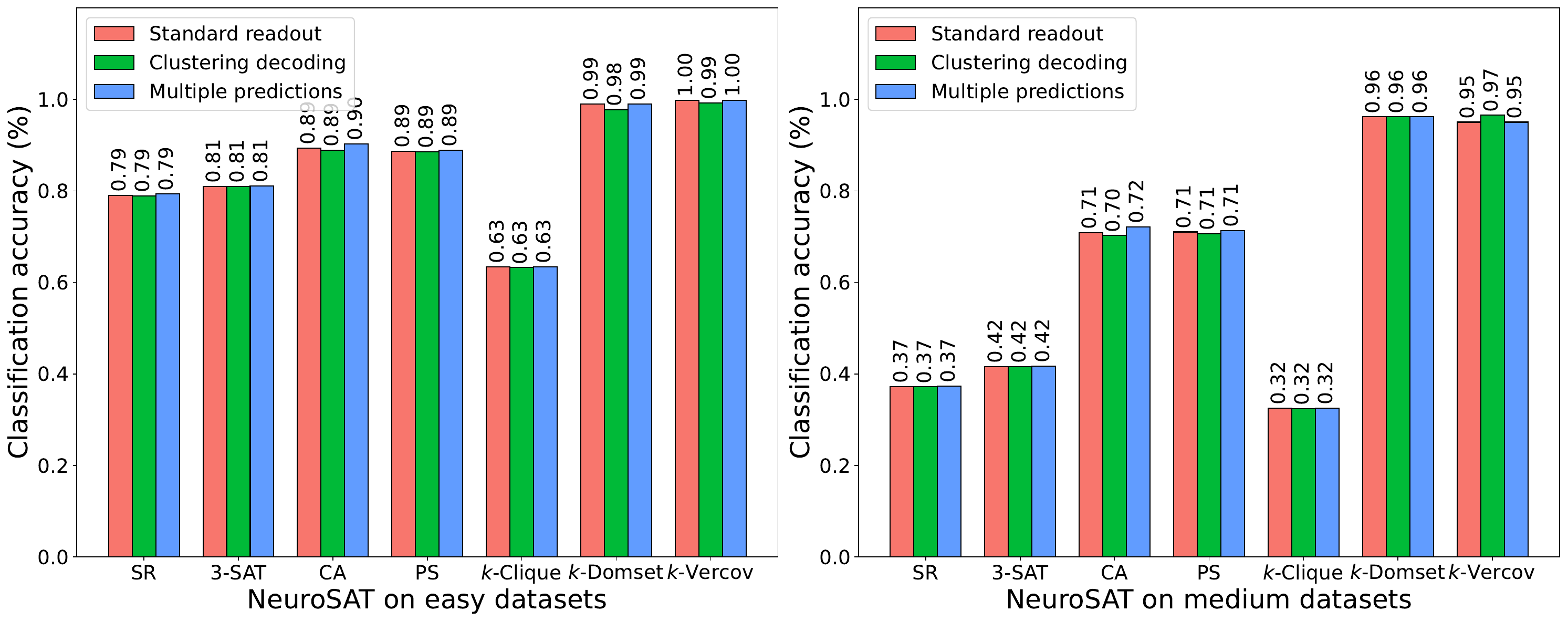}
\caption{Solving accuracy of NeuroSAT with different inference algorithms.}
\label{fig:assignment_decoding}
\end{center}
\vspace{-5pt}
\end{wrapfigure}
Figure~\ref{fig:assignment_decoding} illustrates the results of NeuroSAT using various decoding algorithms (with UNS$_2$ as the training loss). Notably, all three decoding algorithms demonstrate similar performances across all datasets. This observation indicates that utilizing the standard readout after message passing is sufficient for predicting a satisfying assignment. Also, the GNN model has successfully learned to identify potential satisfying assignments within the latent space, which can be extracted by clustering the literal embeddings.

\paragraph{Evaluation with unsatisfiable training instances.}
Following previous works~\citep{circuitsat, pdp, querysat}, our evaluation of GNN models focuses solely on satisfiable instances. However, in practical scenarios, the satisfiability of instances may not be known before training. To address this gap, we explore the effectiveness of training NeuroSAT using the unsupervised loss UNS$_2$ on noisy datasets that contain unsatisfiable instances. Table~\ref{tab:assignment_noisy} presents the results of NeuroSAT when trained on such datasets, where 50\% of the instances are unsatisfiable. Interestingly, incorporating unsatisfiable instances for training does not significantly affect the performance of the GNN model. This finding highlights the potential utility of training GNN models using UNS$_2$ loss on new datasets, irrespective of any prior knowledge regarding their satisfiability.

\begin{table}[h]
  \caption{Solving accuracy of NeuroSAT when trained on noisy datasets. Values in parentheses indicate the performance difference compared to the model trained without unsatisfiable instances. The $k$-Clique dataset is excluded as NeuroSAT fails during training.}
  \label{tab:assignment_noisy}
  \centering
  \resizebox{\textwidth}{!}{
  \begin{tabular}{cccccccccccccc}
    \toprule[1pt]
    \multicolumn{6}{c}{\bf Easy Datasets} & \multicolumn{6}{c}{\bf Medium Datasets} \\
    \cmidrule(lr){1-6}
    \cmidrule(lr){7-12}
     \bf SR & \bf 3-SAT & \bf CA & \bf PS & \bf $k$-Domset & \bf $k$-Vercov & \bf SR & \bf 3-SAT & \bf CA & \bf PS & \bf $k$-Domset & \bf $k$-Vercov \\
    \midrule
    78.84 & 80.48 & 87.01 & 88.66 & 98.00 & 95.24 & 37.21 & 41.75 & 76.49 & 72.52 & 94.93 & 96.18 \\
    (-0.95) & (-0.11) & (-2.33) & (-0.13) & (-0.85) & (-4.49) & (-0.04) & (+0.14) & (+5.64) & (+1.46) & (-1.25) & (+0.19) \\
    \bottomrule[1pt]
  \end{tabular}
  }
\end{table}

\subsection{Unsat-core Variable Prediction}
\label{app:ben_unsat}
\begin{figure}[ht]
    \centering
    \includegraphics[width=\textwidth]{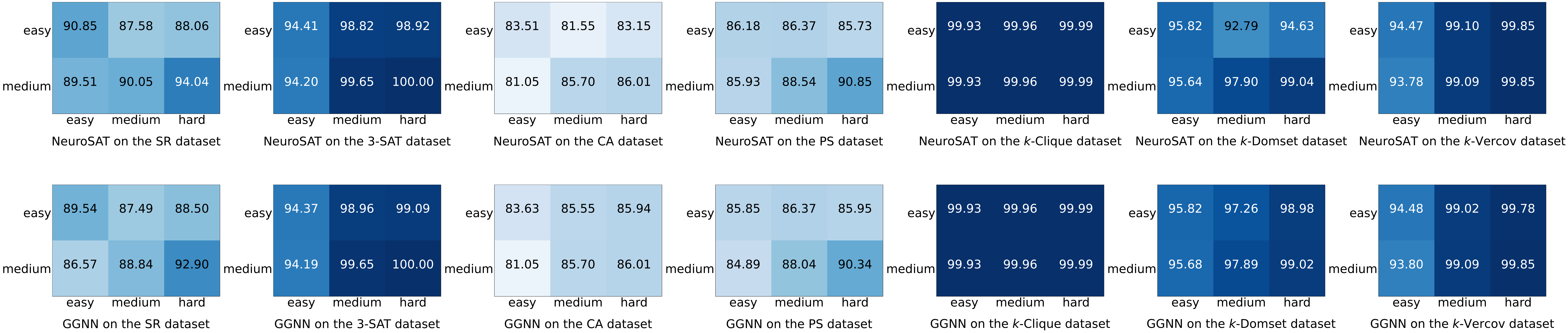}
    \caption{Classification accuracy of unsat-core variables across different difficulty levels. The x-axis denotes testing datasets and the y-axis denotes training datasets.}
    \label{fig:core_difficulty_transfer}
\end{figure}

\paragraph{Evaluation across different difficulty levels.}
The results across different difficulty levels are presented in Figure~\ref{fig:core_difficulty_transfer}. Remarkably, both NeuroSAT and GGNN exhibit a strong generalization ability when trained on easy or medium datasets. This suggests that GNN models can effectively learn and generalize from the characteristics and patterns present in these datasets, enabling them to perform well on a wide range of problem complexities.

\begin{figure*}[ht]
    \centering
    \includegraphics[width=\textwidth]{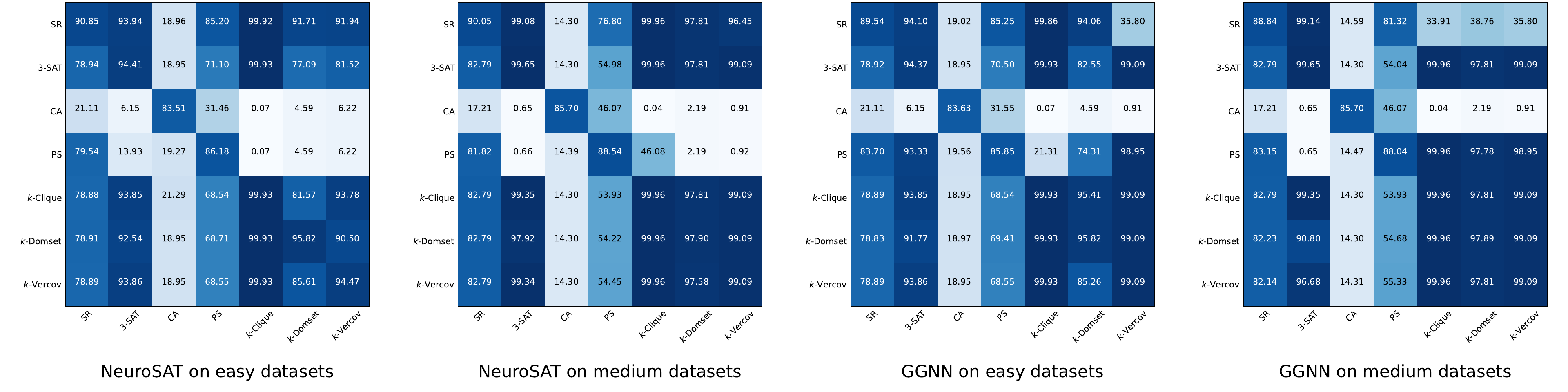}
    \caption{Classification accuracy of unsat-core variables across different datasets. The x-axis denotes testing datasets and the y-axis denotes training datasets.}
    \label{fig:unsat_dataset_transfer}
\end{figure*}

\paragraph{Evaluation across different datasets.}
Figure~\ref{fig:unsat_dataset_transfer} shows the generalization results across different datasets. Both NeuroSAT and GGNN demonstrate good generalization performance to datasets that are different from their training data, except for the CA dataset. This discrepancy can be attributed to the specific characteristics of the CA dataset, where the number of unsat-core variables is significantly smaller compared to the number of variables not in the unsat core. In contrast, other datasets have a different distribution, where the number of unsat-core variables is much larger. This variation in distribution presents a challenge for the models' generalization ability on the CA dataset.

\section{Advancing Evaluation}
\subsection{Implementation details}
\label{app:adv_exp}
To create the augmented datasets, we leverage CaDiCaL~\citep{cadical} to generate a DART proof~\citep{drat} for each SAT instance, which tracks the clause learning procedure and records all the learned clauses during the solving process. These learned clauses are then added to each instance, with a maximum limit of 1,000 clauses. For experiments on augmented datasets, we keep all training settings identical to those used for the original datasets.

For contrastive pretraining experiments, we treat each original formula and its augmented counterpart as a positive pair and all other instances in a mini-batch as negative pairs. We use an MLP projection to map the graph embedding $z_i$ of each formula to $m_i$ and employ the SimCLR’s contrastive loss~\citep{simclr}, where the loss function for a positive pair of examples $(i,j)$ in a mini-batch of size $2N$ is defined as:
\begin{equation}
\mathcal{L}_{i,j} = -\log\frac{\exp(\text{sim}(m_i, m_j)/\tau)}{\sum_{k=1}^{2N}\mathbbm{1}_{[k\neq i]}\exp(\text{sim}(m_i, m_k)/\tau)}.
\end{equation}
Here, $\mathbbm{1}_{[k\neq i]}$ is an indicator function that evaluates to 1 if $k \neq i$, $\tau$ is a temperature parameter, and $\text{sim}(\cdot, \cdot)$ is the similarity function defined as $\text{sim}(m_i, m_j) = m_i^\top m_j / \|m_i\|\|m_j\|$. The final loss is the average over all positive pairs. In our experiments, we set the temperature parameter to 0.5 and utilize a learning rate of $10^{-4}$ with a weight decay of $10^{-8}$. The pretraining process is performed for a total of 100 epochs. Once the pretraining is completed, we keep the GNN model and remove the projection head for downstream tasks.

For experiments involving random initialization, we utilize Kaiming Initialization~\citep{kaiming_initialziation} to initialize all literal/variable and clause embeddings during both training and testing. For the predicted assignments, we utilize 2-clustering decoding to construct two possible assignment predictions for NeuroSAT* at each iteration. When calculating the number of flipped variables and unsatisfiable clauses for NeuroSAT*, we only consider the better assignment prediction of the two at each iteration, which is the one that satisfies more clauses. All other experimental settings remain the same as in the benchmarking evaluation.

\subsection{Comparisons with State-of-the-art SAT Solvers}
\label{app:adv_comparisons}
We compare NeuroSAT with two advanced CDCL and LS solvers, CaDiCaL~\cite{cadical} and Sparrow~\citep{sparrow}. To enable a fair comparison, we first configure Sparrow to generate the same number of assignments as NeuroSAT by setting its maximum flip number to 32, allowing for an apples-to-apples comparison of both solvers' accuracy and execution time. Subsequently, we allow Sparrow and CaDiCaL to run without constraints to solve the satisfiable instances in G4SATBench. Considering that NeuroSAT processes a batch of problems in parallel on GPUs, we calculate its per-instance runtime by dividing the total execution time by the number of testing instances.

The results, summarized in Table~\ref{tab:sparrow}, indicate that GNN-based heuristics could outperform modern local search solvers like Sparrow, generating more satisfying assignments extremely fast when constrained to output a limited number of solutions. However, once such a constraint is lifted, both Sparrow and CaDiCaL can traverse the solution space efficiently and solve all satisfiable instances in \model{}, while GNN models like NeuroSAT may find it challenging due to their limited exploration capacity as evidenced in Figure~\ref{fig:distinct_assignments}. Nevertheless, it's crucial to recognize that while GNN models are hard to compete with CaDiCaL and Sparrow, their assignment predictions could still serve as good initializations in these solvers, potentially leading to better performance~\citep{nlocalsat,nsnet}.

\begin{table}[h]
  \caption{Results of NeuroSAT, Sparrow, and CaDiCaL. The top 2 rows represent the solving accuracy (\%), and the bottom 4 rows represent the running time (second) per instance. Sparrow$^{*}$ refers to Sparrow limited to a maximum of 32 flips.}
  \label{tab:sparrow}
  \centering
  \resizebox{\textwidth}{!}{
  \begin{tabular}{lccccccccccccccc}
    \toprule[1pt]
    \multirow{2}{*}{\bf Method} & \multicolumn{7}{c}{\bf Easy Datasets} & \multicolumn{7}{c}{\bf Medium Datasets} \\
    \cmidrule(lr){2-8}
    \cmidrule(lr){9-15}
     & \bf SR & \bf 3-SAT & \bf CA & \bf PS & \bf $k$-Clique & \bf $k$-Domset & \bf $k$-Vercov & \bf SR & \bf 3-SAT & \bf CA & \bf PS & \bf $k$-Clique & \bf $k$-Domset & \bf $k$-Vercov \\
    \midrule
    NeuroSAT & 79.79 & 80.59 & 89.34 & 88.79 & 63.43 & 98.85 & 99.73 & 37.25 & 41.461 & 70.83 & 71.03 & 32.48 & 96.18 & 95.99 \\
    Sparrow$^{*}$ & 56.03 & 52.09 & 85.48 & 77.25 & 53.68 & 37.15 & 31.61 & 1.64& 1.85 & 12.68 & 10.72 & 12.99 & 1.27 & 0.53 \\
    \midrule\midrule
    NeuroSAT & 0.002 & 0.002 & 0.002 & 0.002 & 0.003 & 0.002 & 0.003 & 0.008 & 0.006 & 0.009 & 0.010 & 0.009 & 0.006 & 0.007 \\
    Sparrow$^{*}$ & 0.005 & 0.005 & 0.006 & 0.006 & 0.005 & 0.005 & 0.005 & 0.008 & 0.007 & 0.006 & 0.006 & 0.006 & 0.007 & 0.006 \\
    \midrule
    Sparrow & 0.007 & 0.007 & 0.008 & 0.008 & 0.007 & 0.009 & 0.009 & 0.013 & 0.013 & 0.010 & 0.013 & 0.013 & 0.012 & 0.011 \\
    CaDiCaL & 0.013 & 0.013 & 0.012 & 0.011 & 0.014 & 0.012 & 0.011 & 0.014 & 0.043 & 0.016 & 0.018 & 0.015 & 0.013 & 0.012 \\
    \bottomrule[1pt]
  \end{tabular}
  }
\end{table}

\end{document}